\documentclass{preprint}

\usepackage{xcolor}
\usepackage{booktabs}
\usepackage{graphicx} 
\usepackage[colorlinks=true]{hyperref}
\usepackage{enumitem}
\usepackage{colortbl}
\usepackage{longtable}
\usepackage{amsmath}
\usepackage{amsfonts}
\usepackage{bbold}
\usepackage{ulem}
\usepackage{makecell}
\usepackage{tabularx}
\usepackage{xltabular}
\usepackage{multirow}

\usepackage[numbers,comma,sort&compress]{natbib}

\usepackage{array}    

\title{Semi-Supervised Learning from Small Annotated Data and Large Unlabeled Data for Fine-grained PICO Entity Recognition}
\author[1,$\dagger$]{Fangyi Chen}
\author[1,$\dagger$]{Gongbo Zhang}
\author[1]{Yilu Fang}
\author[2,*]{Yifan Peng}
\author[1,*]{Chunhua Weng}
\affil[1]{Department of Biomedical Informatics, Columbia University, New York, NY, USA}
\affil[2]{Department of Population Health Sciences, Weill Cornell Medicine, New York, NY, USA}
\affil[*]{Corresponding author(s). Email(s): \url{yip4002@med.cornell.edu}, \url{chunhua@columbia.edu}}
\affil[$\dagger$]{These authors contributed equally to this work.}

\begin{document}

\maketitle

\begin{abstract}
\textbf{Objective:} Extracting PICO elements---Participants, Intervention, Comparison, and Outcomes---from clinical trial literature is essential for clinical evidence retrieval, appraisal, and synthesis. Existing approaches do not distinguish the attributes of PICO entities. This study aims to develop a named entity recognition (NER) model to extract PICO entities with fine granularities.

\textbf{Materials and Methods:} Using a corpus of 2,511 abstracts with PICO mentions from 4 public datasets, we developed a semi-supervised method to facilitate the training of a NER model, FinePICO, by combining limited annotated data of PICO entities and abundant unlabeled data. For evaluation, we divided the entire dataset into two subsets: a smaller group with annotations and a larger group without annotations. We then established the theoretical lower and upper performance bounds based on the performance of supervised learning models trained solely on the small, annotated subset and on the entire set with complete annotations, respectively. Finally, we evaluated FinePICO on both the smaller annotated subset and the larger, initially unannotated subset. We measured the performance of FinePICO using precision, recall, and F1.

\textbf{Results:} Our method achieved precision/recall/F1 of 0.567/0.636/0.60, respectively, using a small set of annotated samples, outperforming the baseline model (F1: 0.437) by more than 16\%. The model demonstrates generalizability to a different PICO framework and to another corpus, which consistently outperforms the benchmark in diverse experimental settings (p-value \textless0.001).

\textbf{Conclusion:} This study contributes a generalizable and effective semi-supervised approach to named entity recognition leveraging large unlabeled data together with small, annotated data. It also initially supports fine-grained PICO extraction.
\end{abstract}


\section{Introduction}

Evidence-based medicine (EBM) has gained increasing popularity over the past decades and has become the guiding principle of medical practice \cite{collins2007evidence, you2016perspective, akobeng2005principles, peng2023ai, zhang2024leveraging}. Aggregating, synthesizing, and understanding the best available clinical evidence is essential to enhancing decision-making in medical practices and optimizing treatment outcomes \cite{berlin2014meta}. Meta-analysis is a crucial statistical technique in evidence synthesis that helps inform the best clinical actions by gathering and combining results from different research studies \cite{cook1997systematic}. However, it is a highly time-consuming and labor-intensive process, making it impractical to constantly keep pace with the rapidly rising number of published studies \cite{berlin2014meta, bastian2010seventy, borah2017analysis}. The PICO (Participants, Intervention, Comparison, and Outcomes) framework serves as the basis for formulating clinical questions and facilitates the efficient retrieval, selection, and categorization of evidence from clinical studies. To enable and streamline the workflow of automated meta-analysis, a more granular PICO characterization is needed for accurately characterizing randomized controlled trials (RCTs). For example, instead of using the category of P to characterize participants, we need more information of different participant attributes such as age, sex, race, and ethnicity.

Automated PICO entity extraction is a named entity recognition (NER) task, wherein each token is tagged with a pre-defined label. Early methods relied on rule-based approaches, Conditional Random Fields (CRF) models, or a combination of basic classifiers \cite{dawes2007identification, demner2007answering, chabou2018combination}. These approaches necessitate exhaustive feature engineering. More recently, the adoption of deep learning algorithms, such as bidirectional long short-term memory (BiLSTM) networks \cite{jin2018pico, graves2005framewise, ma2016end} and BiLSTM models augmented with a CRF module \cite{jin2020advancing, brockmeier2019improving}, have demonstrated superior performance without laborious feature extraction. Later, transformer-based models (e.g., BERT and its variants) have further advanced the field \cite{devlin2018bert, lee2020biobert, liu2019roberta, beltagy2019scibert}.

Despite these advancements, several widely acknowledged challenges persist. One primary challenge is the lack of large, high-quality annotated datasets since annotation is a labor-intensive and time-consuming task that often requires domain experts. Furthermore, the absence of standardized PICO annotation guidelines, which becomes impractical due to variations in study purposes and domains, has further complicated the annotation process. The largest publicly available corpus, EBM-NLP \cite{nye2018corpus}, was reported to exhibit significant inconsistency in annotated results \cite{hu2023towards, lee2019study, abaho2019correcting}. These inconsistencies are mainly attributed to unclear definitions of text span boundaries and complex annotation guidelines, resulting in suboptimal model performances \cite{hu2023towards, lee2019study, abaho2019correcting}. To address these limitations, manual corrections or heuristic rule-based approaches have been leveraged to relabel entities \cite{hu2023towards, abaho2019correcting, dhrangadhariya2023not}. Notably, Hu et al. proposed a two-step NLP pipeline that first classifies sections of sentences and then extracts PICO from sentences in Title and Method sections using BiomedBERT trained on re-annotated abstract \cite{hu2023towards}. Although their proposed method reduced annotation time for sentences rich in PICO information and achieved high inter-annotator agreement, the overall number of annotated abstracts remained considerably limited.

Another issue is the lack of fine-grained annotation. Most public datasets only provide coarse-level PICO annotations \cite{sanchez2022annotated}, which do not always meet the requirements for many downstream tasks, such as meta-analysis or evidence appraisal. Although the EBM-NLP dataset was unusually annotated with fine-grained PICO entities, these annotations are unsuitable for meta-analysis because they do not capture numeric values associated with outcome measures for different study arms (e.g., intervention and control). The ability to extract numerical data is critical for conducting a statistical analysis to evaluate the efficacy of the intervention \cite{ahn2018introduction}. Nevertheless, limited effort has been dedicated to extracting detailed outcome information, \emph{e.g.,} the number of subjects experiencing specific outcome events. Mutinda et al. introduced a fully annotated dataset comprising 1,011 randomized controlled trials (RCTs) on breast cancer \cite{mutinda2022pico}. While their PICO annotation framework was suitable for conducting meta-analysis, it did not include annotations for key population characteristics (e.g., sex) because their selected RCTs focused mainly on the female population. Therefore, the generalizability of NER models built using this dataset was significantly compromised.

Recognizing these challenges, we proposed FinePICO, a semi-supervised learning (SSL) algorithm to enhance the extraction of fine-grained PICO entities. SSL is a branch of machine learning model that utilizes both labeled and unlabeled data for model training \cite{adeva2014automatic}. Compared to fully supervised learning that demands a vast number of labeled samples to achieve optimal performance, SSL effectively leverages abundant unlabeled data combined with scarce labeled data to improve learning outcomes. Current PICO extraction models heavily depend on the availability and quality of annotated samples, which are challenging to obtain and inconsistent across sites, thereby limiting their robustness and generalizability \cite{jin2018pico, brockmeier2019improving, nye2018corpus, hu2023towards}. In contrast, SSL offers significant advantages in low-resource settings where labeled data is expensive and sparse. While several limitations have been acknowledged, such as higher computational costs, risk of propagating errors, and assumption about data distribution \cite{yang2022survey, yang2023prototype}, SSL has been widely adopted and demonstrated promising results in various applications, such as object recognition and image segmentation \cite{hong2015decoupled, roli2006semi}, document retrieval and classification \cite{yang2023prototype, bickel2006ecml, li2009semi, duh2011semi}, and biomedical information mining \cite{erkan2007semi}. The primary focus of this study is to explore SSL in fine-grained PICO extraction, as its efficacy in this area remains uncertain.''

Our main objective was to demonstrate that combining limited labeled data and a substantial volume of unlabeled data can achieve performance comparable to that of models trained using fully annotated data. Our findings suggested that SSL techniques can optimize fine-grained PICO extraction by greatly expanding the training sample size while minimizing reliance on extensive manual annotation efforts.

\section{Materials and Methods}

\subsection{Workflow Overview}\label{workflow-overview}

\begin{figure}
    \centering
    \includegraphics[width=\linewidth]{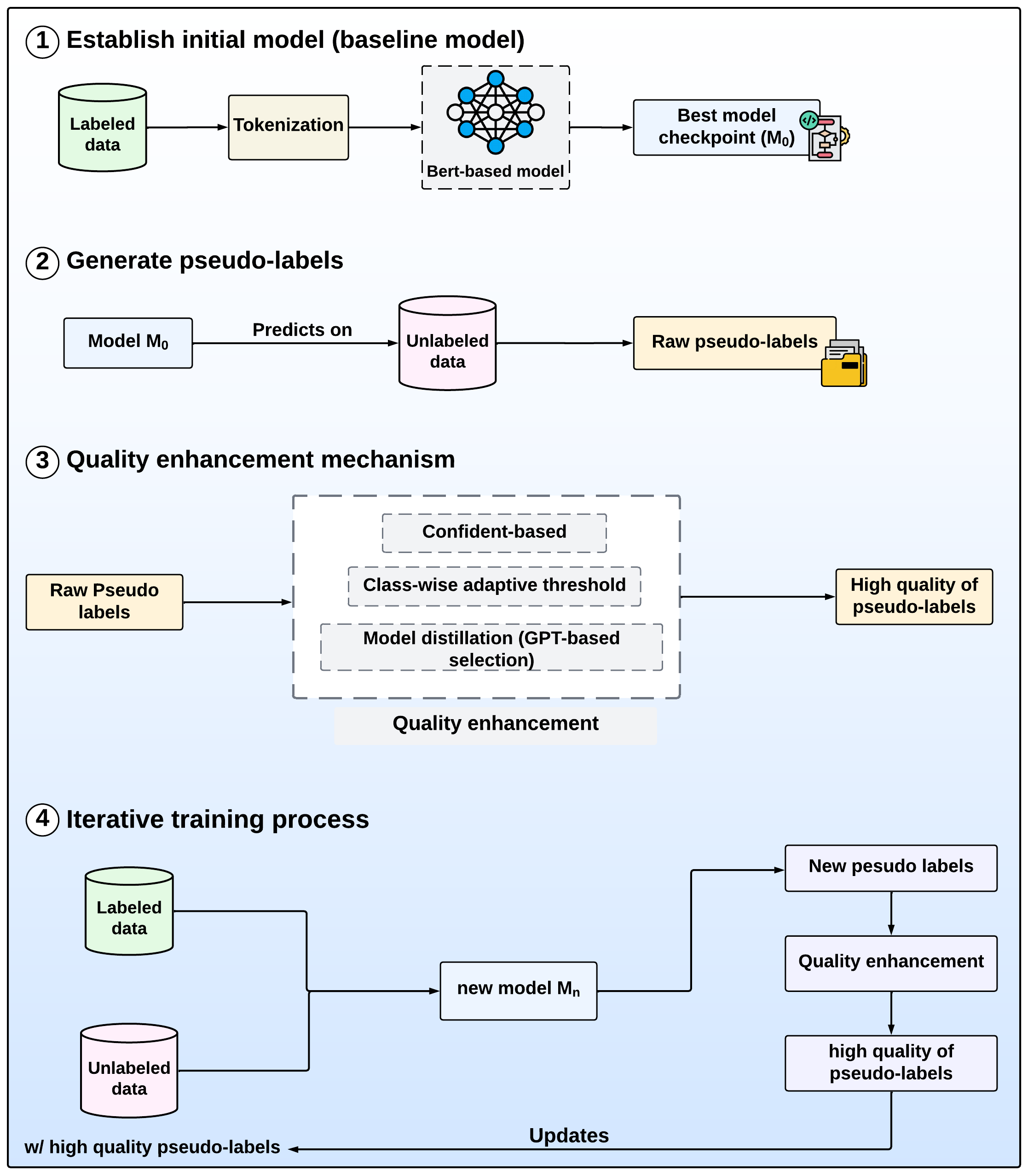}
    \caption{The overview of the study workflow.}
    \label{fig:workflow}
\end{figure}

FinePICO employs an iterative SSL process to adjust model weights and generate pseudo-labels for unlabeled data. Figure \ref{fig:workflow} depicts the overview design of the model. Specifically, we first develop a NER model using the available annotated data via the traditional supervised learning approach. Once the initial model is trained, it is deployed to make inferences on the unlabeled data, referred to as pseudo-labels. We enrich the original labeled data with the high-confidence pseudo-labeled data for fine-tuning the model. We iteratively repeat the cycle of generating pseudo-labels and updating model weights until the model's performance converges on the validation dataset or a predefined maximum number of iterations has been reached. To ensure the quality of the pseudo-labels and minimize the risk of error propagation, we incorporate a quality enhancement module. It performs a quality check and selects the generated labels with high confidence.

\subsubsection{Foundation Model}\label{foundation-model}

To leverage the power of pre-trained language models, we select a BERT-based model as our foundation model \cite{gu2021domain}. We define \(S\) as the entire collection of sentences of interest, where \(S_{label}\) refers to the sentences with pre-annotated named entity tags associated with their tokens. For each sentence \(s_{i}^{l} \in S_{label}\), we have a sequence of tokens \(\left\{ t_{i1}^{l}, {t}_{i2}^{l},\ldots,t_{im}^{l} \right\}\), where each token \(t_{ij}^{l}\) is associated with a label \(y_{ij}^{l}\), and \(m\) is the length of the sentence \(s_{i}^{l}\).

We also define \(S_{unlabel}\ \)as the set of sentences without annotated named entity tags. We leverage the BERT-based model that was previously trained on \(S_{label}\) to make inferences on \(S_{unlabel\ }\) and generate the set of pseudo-labels (\({\hat{y}}_{ij}^{u}\)) for each token in the unlabeled sentence \(s_{i}^{u} \in S_{unlabel}\).

The training and fine-tuning process involves applying the softmax function \(\sigma(.)\) on the last layer of the neural network to compute the probability \(p_{ij}^{k}\) for the \(k\)\textsuperscript{th} entity class associated with the token \(t_{ij}\). The predicted entity class \({\hat{y}}_{ij}\) is then determined as follows:
\begin{align}
p_{ij}^{k} &= \sigma(\mathbf{z})_{ij}^{k} = \frac{\exp(z_{ij}^{k})}{\sum_{v=1}^{C}{\exp(z_{ij}^{v})}}\\
\hat{y}_{ij} & = \arg\max\left(\sigma(\mathbf{z})_{ij}\right),\ p_{ij}^{k} \subseteq \sigma(\mathbf{z})_{ij}
\end{align}
where \(\mathbf{z}\) is an embedding-based representation of each token, and \(C\) is the total number of entity class. \(\sigma\left( \mathbf{z} \right)_{ij}\) represents probabilities across entity tags for token \(t_{ij}\). The target function is to minimize the cross-entropy loss function. The loss function at token \(t_{ij}\) is defined below:
\begin{equation}
\mathcal{L}_{CE_{ij}} = - \sum_{k=1}^{C}\mathbb{1}(y_{ij}=k)\log p_{ij}^{k}
\end{equation}
The binary indicator \(\mathbb{1}(*)\in \{0,1\}\) equals to 1 if a token belongs to the \(k^{th}\) class and 0 otherwise. The overall loss function comprises of two parts: the supervised loss (\(\mathcal{L}_{s}\)) and unsupervised loss (\(\mathcal{L}_{u}\)).
\begin{equation}
\mathcal{L}_{total} = \mathcal{L}_{s} + \alpha\mathcal{L}_{u}
\end{equation}

\subsubsection{Supervised Learning Loss}\label{supervised-learning-loss}

We leveraged \(S_{label}\) as the main dataset for training and developing our initial baseline models \(M_{0}\). The training process follows well-established supervised learning methods. In this stage, we aim to develop a model that can make reasonable inferences on unseen data. The baseline models were then iteratively refined using both \(S_{label}\) and \(S_{unlabel}\) to minimize the learning loss. The total supervised learning loss \(\mathcal{L}_{s}\) at \(t^{th}\) iteration is computed as follows:
\begin{equation}
\mathcal{L}_{s} = - \frac{1}{\sum_{q=1}^{n^{l}}{m_{q}^{l}}}\sum_{i}^{n^{l}}{\sum_{j}^{m^{l}}\mathcal{L}_{{CE}_{ij}}^{s}}
\end{equation}
where \(n^{l}\) refers to the number of sentences with annotation and \(m^{l}\ \)is the number of tokens at \(i^{th}\) sentence. \(\mathcal{L}_{{CE}_{ij}}^{s}\)denotates as the supervised learning loss function at token \(t_{ij}\).

\subsubsection{The Quality enhancement Mechanism of Pseudo-label Generation}\label{the-quality-enhancement-mechanism-of-pseudo-label-generation}

The baseline model \(M_{0}\) infers labels for each token in the unlabeled sentences. We incorporated the sets of pseudo-labels \(\left\{ {\hat{y}}_{i1}^{u},\hat{y}_{i2}^{u},\ldots,\hat{y}_{im}^{u} \right\}\) with \(\left\{ t_{i1}^{u},{t}_{i2}^{u},\ldots,t_{im}^{u} \right\}\) of the sentence \(s_{i}^{u} \in S_{unlabel}\) into the original training pool \(S_{label}\) to further improve \(M_{0}\). For a token \(t_{ij}^{u}\) in the sentence \(s_{i}^{u}\), its pseudo-label is formally defined as:
\begin{equation}
\hat{y}_{ij}^{u} = \arg\max\left( \sigma(\mathbf{z}^{u})_{ij} \right)
\end{equation}

To maintain the quality and consistency of the generated pseudo-labels on a diverse set of training samples, we introduced a quality enhancement module to select the high-quality labels that would be used in subsequent training iterations. Specifically, we implemented three different quality enhancement approaches within the label selection process and evaluated their relative effectiveness in enhancing the overall model performances.

The selective unsupervised learning loss of a token is computed as follows:
\begin{align}
\mathcal{L}^{u}_{CE_{ij}} &= - \sum_{k=1}^{C}\mathbb{1}( {\hat{y}}_{ij}^{u})\log p_{ij}^{k}\\
\mathbb{1}(\hat{y}_{ij}^{u}) &= \mathbb{1}({\hat{y}}_{ij}^{u} = k )\land \mathbb{1}\left( f(\hat{y}_{ij}^{u},t_{ij}^{u}) \right)
\end{align}
where the binary indicator \(\mathbb{1}(\hat{y}_{ij}^{u}) = 1\)when the two conditions are met simultaneously. The quality enhancement function \(f\) minimizes noises resulting from erroneous predictions by checking if the pseudo-label \({\hat{y}}_{ij}^{u}\) is accurate or has a high degree of certainty. In this study, we investigated three checking strategies.

\begin{enumerate}[nosep]
\def\labelenumi{\arabic{enumi})}
\item
  \textbf{Confident-based masking}. This approach leverages prior studies that revealed the benefits of masking out low-confident examples from the training set \cite{ferreira2023using, xie2020unsupervised}. It uses a predefined threshold to filter out pseudo-labels lower than this level. The threshold is empirically determined to balance between maintaining high label quality and retaining a sufficient volume of training samples.
\item
  \textbf{Class adaptive threshold-based masking}. A recognized limitation of confident-based masking is its potential bias toward classes with higher quality pseudo-labels \cite{zhang2021biasing}. To address this issue, we also implemented a class-wise threshold adjustment algorithm, where the threshold for entity class \(k\) is dynamically calculated per iteration:
\begin{equation}
\tau_{k} = \frac{\sum_{i= 1}^{n^{u}}{\max_{j}{P(k|t_{ij}^{u})}}}{\sum_{i=1}^{n^{u}}{\sum_{j= 1}^{m^{u}}{\mathbb{1}(\hat{y}_{ij}^{u} = k)}}}
\end{equation} 
where \(n^{u}\) denotes the number of unlabeled sentences and \(m^{u}\) refers to the number of unlabeled tokens. We update the threshold for each class and filter the token and its label if the associated probability is less than the dynamic threshold \(\tau_{k}\).
\item
  \textbf{Label Selection via Model Distillation (GPT-based Selection)}. We leverage GPT-4o to evaluate the pseudo-label quality. With the tokenized sentences as input, we prompt GPT-4o to confirm whether the pseudo-labels are correct. Inspired by \citet{hu2024improving}, we curate customized prompts for different entities. Each prompt includes annotation guidelines, error-based instruction, as well as a few annotated examples (Supplementary Table \ref{sup tab:prompt}) The labels confirmed as accurate by GPT-4o are then incorporated into the new training dataset.
\end{enumerate}

\subsection{Data Source}\label{data-source}

We tested FinePICO with different data augmentation strategies, including the use of in-domain data, cross-domain data, and both. In-domain augmentation refers to the scenario where the labeled and unlabeled data are sampled from the same domain, while cross-domain augmentation refers to the scenario where the labeled and unlabeled are sampled from different domains.

For this purpose, we used four public datasets in this study, including PICO-Corpus \cite{mutinda2022pico}, EBM-NLP \cite{nye2018corpus} samples (n = 1,200 abstracts), and two sets of RCT abstracts \cite{hu2023towards} focused on Alzheimer's disease (AD) and COVID-19. The number of PICO entities is summarized in Table \ref{tab:Characteristics}.
\begin{table}
\centering
\caption{Characteristics for four datasets used in this study. INT - intervention arm. CTL - control group.}
\label{tab:Characteristics}
\small
\begin{tabular}{lrrrr}
\toprule
& \textbf{PICO-Corpus} & \textbf{EBM-NLP} & \textbf{AD} & \textbf{COVID-19} \\
\midrule
\textbf{Abstracts} & \textbf{1,011} & \textbf{1,200} & \textbf{150} & \textbf{150} \\
Training & 1010 & & & \\
Validation & 645 & & & \\
Test & 944 & & & \\
\midrule
\textbf{Population (P)} & & 3,951 & 215 & 262 \\
Total sample size & 1,094 & - & - & - \\
Sample size in INT & 887 & & & \\
Sample size in CTL & 784 & - & - & - \\
Age & 231 & - & - & - \\
Eligibility & 925 & - & - & - \\
Ethnicity & 101 & - & - & - \\
Condition & 327 & - & - & - \\
Location & 186 & - & - & - \\
\midrule
\textbf{Intervention (I)} & 1,067 & 5,916 & 467 & 602 \\
\midrule
\textbf{Control (C)} & 979 & 563 & 103 & 180 \\
\midrule
\textbf{Outcome (O)} & & 7,151 & 626 & 626 \\
Study outcomes & 5,053 & - & - & - \\
Outcome measures & 1,081 & - & - & - \\
Binary outcomes & & & & \\
- Absolute value, INT/CTL & 556/465 & - & - & - \\
- Percentage values, INT/CTL & 1,376/1,148 & - & - & - \\
Continuous outcomes & & & & \\
- Mean, INT/CTL & 336/327 & - & - & - \\
- Median, INT/CTL & 270/247 & - & - & - \\
- Standard deviation, INT/CTL & 129/124 & - & - & - \\
- q1, INT/CTL & 4/4 & - & - & - \\
- q3, INT/CTL & 4/4 & - & - & - \\
\bottomrule
\end{tabular}
\end{table}

PICO-Corpus includes 1,011 RCTs related to breast cancer, where each abstract was manually annotated for the pre-defined PICO subcategories (e.g., total sample size, age, and outcome values) \cite{mutinda2022pico}. EBM-NLP corpus composes RCT abstracts in diverse domains, where the training set of the abstracts was annotated by Amazon Mechanical Turk and inter-annotator conflicts were resolved via a voting strategy. Previous studies reported a lack of consistency and agreement among the annotators \cite{nye2018corpus, lee2019study, abaho2019correcting}, with Cohen's kappa coefficient of inter-rater reliability being 0.3 \cite{hu2023towards}. Due to these limitations, we adopted the annotation scheme in PICO-Corpus and utilized EBM-NLP mainly for training data augmentation. We randomly picked 1,200 abstracts from EBM-NLP. The two datasets of AD and COVID-19 did not provide fine-grained PICO annotation; as such, these two were reserved for testing purposes only.

Following the preprocessing workflow of earlier studies \cite{zhang2024span, mutinda2022automatic}, we extracted PICO entities from each sentence in the abstract. The RCT abstracts (n = 2,511) were tokenized into sentences using a Python library NLTK \cite{bird2009natural}. We divided sentences from PICO-Corpus into training, validation, and testing sets. The train-test splitting ratio was set to 80:20, and within the training set, we reserved 10\% of sentences for validation. Clinical trials in EBM-NLP with PICO annotations removed were included as the unlabeled data in the training set. The two datasets, AD and COVID-19, were reserved for testing purposes. We adopted the BIO2 tagging schema in this task \cite{sang2000introduction, he2023document}, which is widely used in NER tasks. Specifically, each token in a sequence is labeled with a combination of a prefix and the type of predefined entities. The prefix indicates the beginning (B), inside (I), or outside (O) of the entities. The common method follows a two-step process that first identifies the relevant entities and then performs relationship extraction to determine intervention and control values. In our study, we streamlined the workflow by eliminating the relationship extraction step, as our detailed annotation labels explicitly classify these values into distinct entity categories (Table~\ref{tab:Characteristics}).

\subsection{Foundation Model Choice \& Baseline Model}\label{foundation-model-choice-baseline-model}

We first tested several open-source models (e.g., BiomedBERT \cite{gu2021domain}, BioBERT \cite{lee2020biobert}, SciBERT \cite{beltagy2019scibert}, ClinicalBERT \cite{wang2023optimized}) used by previous studies to extract fine-grained PICO entities. These models were built using all the labeled training data and were evaluated on the test set. We followed the same hyper-parameter settings described in the prior works \cite{hu2023towards, mutinda2022automatic}, using a learning rate of 5e-5, a batch size of 8, and a total of 10 training epochs.

The performances of several BERT-based models (BioBERT, SciBERT, ClinicalBERT, BiomedBERT) are detailed in Supplementary Table \ref{sup tab:performance}. BiomedBERT achieved the highest macro-average precision of 0.662, recall of 0.716, and F1 score of 0.688 in extracting fine-grained PICO elements, outperforming the other models. Such results aligned with the findings of a previous study focusing on extracting granular PICO information from texts \cite{zhang2024span}, suggesting the superior performance of BiomedBERT in identifying PICO entities. Therefore, in the remaining experiments, we used BiomedBERT as the baseline model.

Considering the constraints of limited available annotations, we defined an ideal scenario where the unlabeled data would be annotated by human experts. We used the model performance from this ideal scenario as the upper bound of SSL model performance in our experiments.

\subsection{Data Augmentation with Unlabeled Data}\label{data-augmentation-with-unlabeled-data}

We augmented the training data with unlabeled text corpus from three distinct domains: in-domain (similar domain with the labeled data), cross-domain (different domains from the labeled data: EBM-NLP), and all-domain (both in-domain and cross-domain unlabeled data). To evaluate the in-domain and all-domain cases, we masked out annotations with different ratios in the training data. Specifically, we randomly selected 10\%, 30\%, 50\%, 70\%, 90\%, and 100\% of the sentences from the training set to act as labeled data and treat the rest as unlabeled data (Supplementary Table \ref{sup tab:scheme}). The proposed algorithm was assessed across these different masking ratios and compared with the performances of the baseline model.

\subsection{Generalizability Test on an Enhanced PICO Scheme}\label{generalizability-test-on-an-enhanced-pico-scheme}

To demonstrate generalizability, we evaluated FinePICO on a newly annotated dataset under a revised guideline adopted from the one used for PICO-Corpus. The first change is a new demographic entity representing the genders of participants. Gender is an important demographic characteristic that enables the exploration of varying treatment effects across different gender subgroups \cite{assmann2000subgroup, bhandari2006misuse}; however, it was not included in the original annotation scheme.

To streamline the gender entity labeling process, we constructed a gender entity tagger using the BiomedBERT fine-tuned on carefully selected samples from EBM-NLP. The samples were selected by first extracting sentences containing tokens tagged with the ``sex'' entity label, followed by manual validation, and supplemented by a keyword search approach to ensure accurate extraction of the sex entity from the text. The final data comprised 569 sentences, partitioned with 80\% for training, 10\% for validation, and 10\% for testing.

We trained the model for 5 epochs with a learning rate of 5e-5, achieving a high F1 score of 0.989. The best-performing model was then utilized to recognize sex entities in the PICO-Corpus (training and validation set). Finally, two researchers (FC, YF) manually annotated the sex entity in the testing set to provide a benchmark, with Cohen's kappa score of 0.98.

The second change involves replacing and consolidating several categories to enhance clarity and efficiency. The revised PICO scheme is illustrated in Figure \ref{fig:scheme}, and the details of the entity counts can be found in Supplementary Table \ref{sup tab:entity_count}. Specifically, we combined the ``subject eligibility'' and ``conditions'' into a single entity group now named ``recruited participant eligibility conditions.'' This merger reflects their interrelated nature and simplifies the tagging process. Additionally, we combined ``outcome names'' and ``outcome measures'' into one group to avoid redundancy and streamline the dataset.

\begin{figure}
    \centering
    \includegraphics[width=\linewidth]{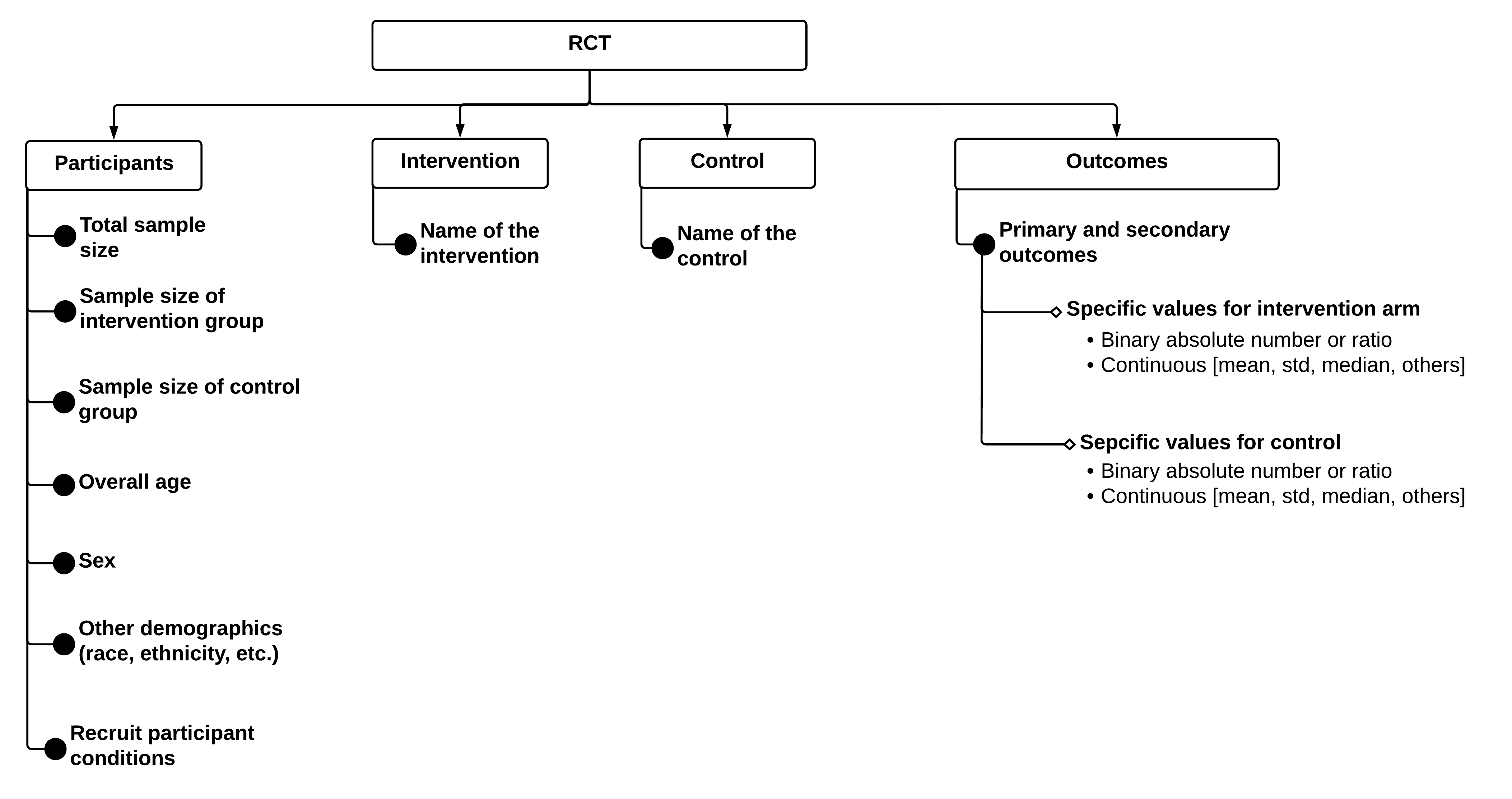}
    \caption{The enhanced PICO scheme.}
    \label{fig:scheme}
\end{figure}

\subsection{Evaluation Metrics}\label{evaluation-metrics}

We tested our models on two independent test sets (PICO-Corpus, AD, and COVID-19 from \citet{hu2023towards}). In the first test set derived from the PICO-Corpus, we evaluated our NER models at a strict entity level that requires the recognition of the complete span of each entity. Since token-level evaluation can be misleadingly high for the intended task, as missing tokens could result in significant misinterpretation, it is essential to accurately capture entire PICO entities. We computed the macro-average precision, recall, and F1 score using seqeval \cite{nakayama2018seqeval}, a well-tested tool often deployed in numerous NLP studies for system evaluation \cite{heddes2021automatic}. The 95\% confidence interval of the performance was estimated based on the bootstrapped test samples.

Acknowledging the variance in annotated spans across different datasets, we conducted a second evaluation using partial-matching \cite{seki2003probabilistic} on AD and COVID-19 datasets. Here, we counted a predicted named entity as a true positive if it overlaps with the human-labeled entities with at least one token. It is worth noting that AD and COVID-19 did not include fine-grained PICO annotation. Therefore, we first converted the predicted fine-grained entities into coarse-level entities and evaluated them using a partial matching strategy \cite{seki2003probabilistic}.

\section{Results}\label{results}

\subsection{Performance on Limited Labeled Samples }{Performance on Limited Labeled Samples}\label{performance-on-limited-labeled-samples}

The baseline models were established solely using labeled samples. The lower bound performance refers to the baseline model evaluated on the test set, whereas the upper bound corresponds to the model trained on the entire set of labeled training samples and evaluated on the test set.

\begin{figure}
    \centering
    \includegraphics[width=\linewidth]{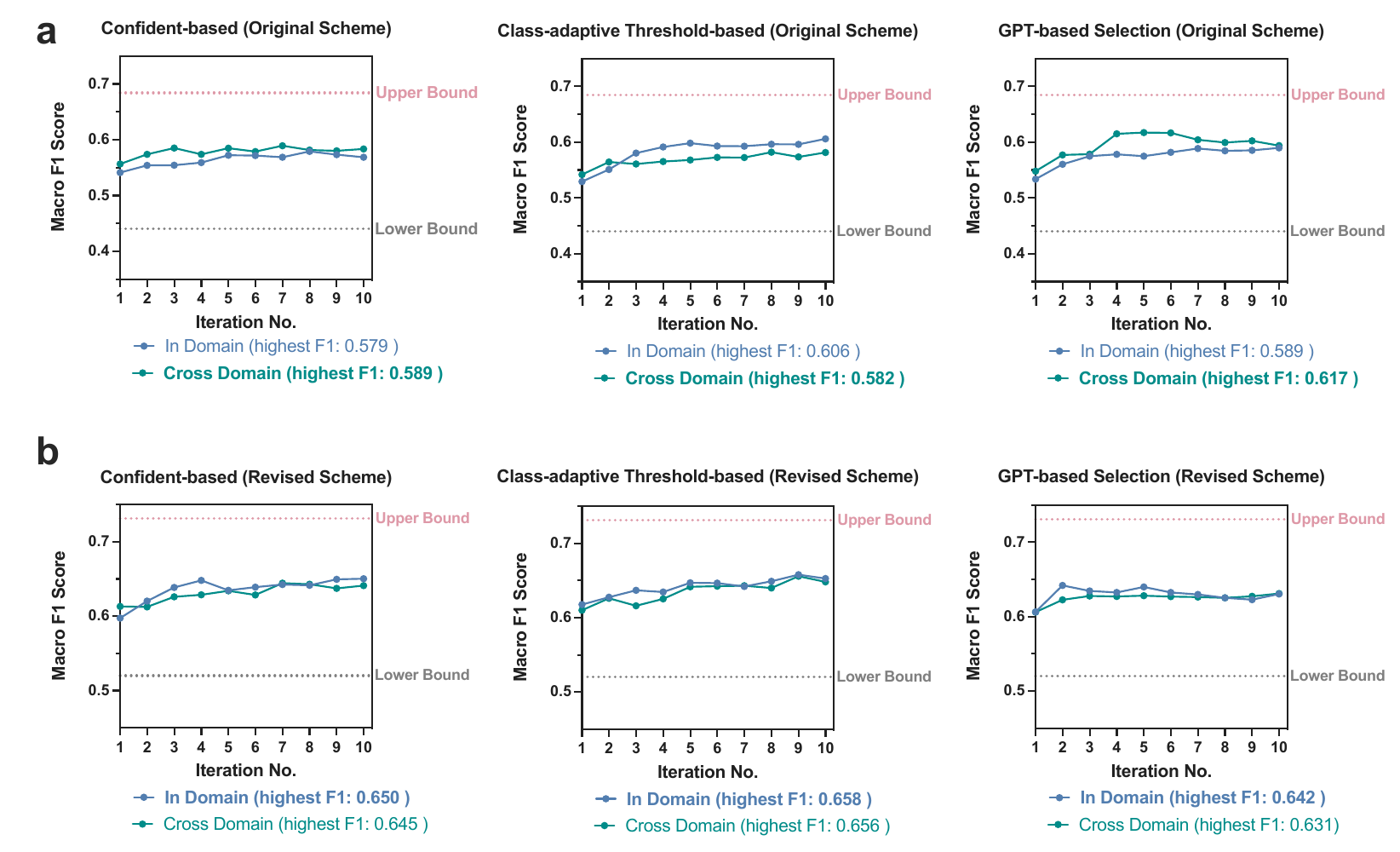}
    \caption{Performance of the proposed models using 10\% annotated data augmented with in-domain, cross-domain data. Lower bound performance is detonated as the baseline model evaluated on the test set. The upper bound refers to the baseline model trained using the whole labeled training samples and evaluated on the test set.}
    \label{fig:performance}
\end{figure}

In scenarios where limited labeled samples were available (e.g., case 1 with 10\% labeled data, as shown in Supplementary Table \ref{sup tab:scheme}), FinePICO notably surpassed the lower bound benchmarks in both data augmentation settings during the iterative training process (Figure \ref{fig:performance}). For instance, employing the confident-based approach, the model augmented with cross-domain data achieved the highest macro-average F1 score of 0.589 at the 7\textsuperscript{th} iteration. This score marked an approximately 15\% improvement over the lower bound (F1-score of 0.44). Similarly, statistical improvements over the baseline model were observed when different data augmentation strategies were applied, and when the model was adapted to the revised PICO scheme.

\subsection{Performance Comparison of Different Quality Enhancement Approaches}\label{performance-comparison-of-different-quality-enhancement-approaches}

The performances of three quality enhancement strategies for optimizing pseudo-label selection are summarized in Table \ref{tab:performance}. All three quality enhancement methods outperformed the baseline models by over 10\% in precisions, recall, and F1 scores, with their respective 95\% confidence intervals (CIs) provided in Supplementary Table \ref{sup tab:clean_perf}. In the original PICO scheme, GPT-based selection achieved the highest performance (average F1 of 0.6, 95\% CI between 0.609 and 0.664) among the three methods. However, we did not perceive any statistical enhancement (p-value =0.171) using GPT-based selection over the confident-based masking algorithm. In the revised PICO scheme, the adaptive threshold-based method was the most effective in selecting high-quality pseudo-labels among the three quality enhancement approaches, obtaining the highest average F1 score of 0.653 (95\% CI: 0.657 - 0.706) when augmented with in-domain unlabeled data. Additionally, both confident-based and adaptive threshold-based masking methods have performed statistically better than GPT-based selection (p-value \textless{} 0.05).

\begin{table}
\centering
\caption{Average performance of different quality enhancement approaches evaluated on the bootstrapped testing samples. R -- Recall. P -- Precision.}
\label{tab:performance}
\small
\begin{tabular}{lrrrrrr}
\toprule
Quality enhancement Approaches & \multicolumn{3}{c}{Original Scheme} &  \multicolumn{3}{c}{Revised Scheme} \\
\cmidrule(rl){2-4}\cmidrule(rl){5-7}
 & R & P & F1 & R & P & F1\\
\midrule
\textbf{Confident-based masking} \\
In-domain & 0.607 & 0.566 & 0.586 & 0.675 & 0.628 & 0.651 \\
Cross-domain & 0.619 & 0.580 & 0.598 & 0.652 & 0.613 & 0.632 \\
\midrule
\textbf{Class adaptive threshold masking}\\
In-domain & 0.636 & 0.561 & 0.596 & 0.682 & 0.626 & 0.653 \\
Cross-domain & 0.617 & 0.571 & 0.594 & 0.677 & 0.627 & 0.651 \\
\midrule
\textbf{GPT-based selection}\\
In-domain & 0.607 & 0.591 & 0.599 & 0.639 & 0.607 & 0.622 \\
Cross-domain & 0.636 & 0.567 & 0.600 & 0.613 & 0.608 & 0.610 \\
\midrule
\textbf{Baseline Model}\\
BiomedBERT & 0.489 & 0.394 & 0.437 & 0.568 & 0.480 & 0.520 \\
\bottomrule
\end{tabular}

\end{table}

\subsection{Generalizability Assessment}\label{generalizability-assessment}

To assess the generalizability of FinePICO, with the consideration of available resources, we selected confident-based masking as the primary quality enhancement approach. The best-performing models were examined on additional data augmentation cases ranging from 30\% to 100\% of annotated samples.

\subsubsection{Additional Data Augmentation Scenarios}\label{additional-data-augmentation-scenarios}

Table \ref{tab:bootstrapped} presents the average performances of models with different data augmentation cases, with the baseline levels detailed in Supplementary Table \ref{sup tab:baseline_comp}. Our analysis revealed a positive linear relationship between model performance and the number of annotated samples used for training. Specifically, performance increased from an F1 score of 0.667 (cross-domain) with 30\% of the annotated data to 0.695 with the entire labeled data. This suggests that while additional labeled data continues to improve the model performance, the marginal gains diminish as the proportion of annotations approaches 100\%.

\begin{table}
\centering
\caption{Average Performances on bootstrapped testing samples. R -- Recall. P -- Precision}
\label{tab:bootstrapped}
\small
\begin{tabular}{lrrrrrr}
\toprule
Data Augmentation Cases & \multicolumn{3}{c}{Original Scheme} & \multicolumn{3}{c}{Revised Scheme} \\
\cmidrule(rl){2-4}\cmidrule(rl){5-7}
 & R & P & F1 & R & P & F1 \\
\midrule
\textbf{Case 2: 30\%   Annotation} &  &  &  &  &  &  \\
In-domain & 0.673 & 0.623 & 0.647 & 0.722 & 0.668 & 0.694 \\
Cross-domain & 0.674 & 0.616 & 0.644 & 0.712 & 0.650 & 0.680 \\
All & 0.689 & 0.645 & 0.667 & 0.708 & 0.675 & 0.691 \\
\midrule
\textbf{Case 3: 50\% Annotation} &  &  &  &  &  &  \\
In-domain & 0.687 & 0.647 & 0.667 & 0.737 & 0.702 & 0.719 \\
Cross-domain & 0.699 & 0.647 & 0.672 & 0.717 & 0.691 & 0.703 \\
All & 0.699 & 0.650 & 0.673 & 0.730 & 0.700 & 0.714 \\
\midrule
\textbf{Case 4: 70\%   Annotation} &  &  &  &  &  &  \\
In-domain & 0.699 & 0.663 & 0.681 & 0.734 & 0.699 & 0.716 \\
Cross-domain & 0.702 & 0.649 & 0.674 & 0.737 & 0.697 & 0.716 \\
All & 0.699 & 0.645 & 0.646 & 0.735 & 0.700 & 0.717 \\
\midrule
\textbf{Case 5: 90\%   Annotation} &  &  &  &  &  &  \\
In-domain & 0.715 & 0.663 & 0.688 & 0.749 & 0.703 & 0.725 \\
Cross-domain & 0.728 & 0.672 & 0.699 & 0.750 & 0.706 & 0.727 \\
All & 0.717 & 0.678 & 0.697 & 0.742 & 0.693 & 0.717 \\
\midrule
\textbf{Case 6: 100\%   Annotation} &  &  &  &  &  &  \\
In-domain & - & - & - & - & - & - \\
Cross-domain & 0.716 & 0.676 & 0.695 & 0.753 & 0.713 & 0.732 \\
All & 0.716 & 0.676 & 0.695 & 0.753 & 0.713 & 0.732 \\
\bottomrule
\end{tabular}
\end{table}

\begin{figure}
    \centering
    \includegraphics[width=\linewidth]{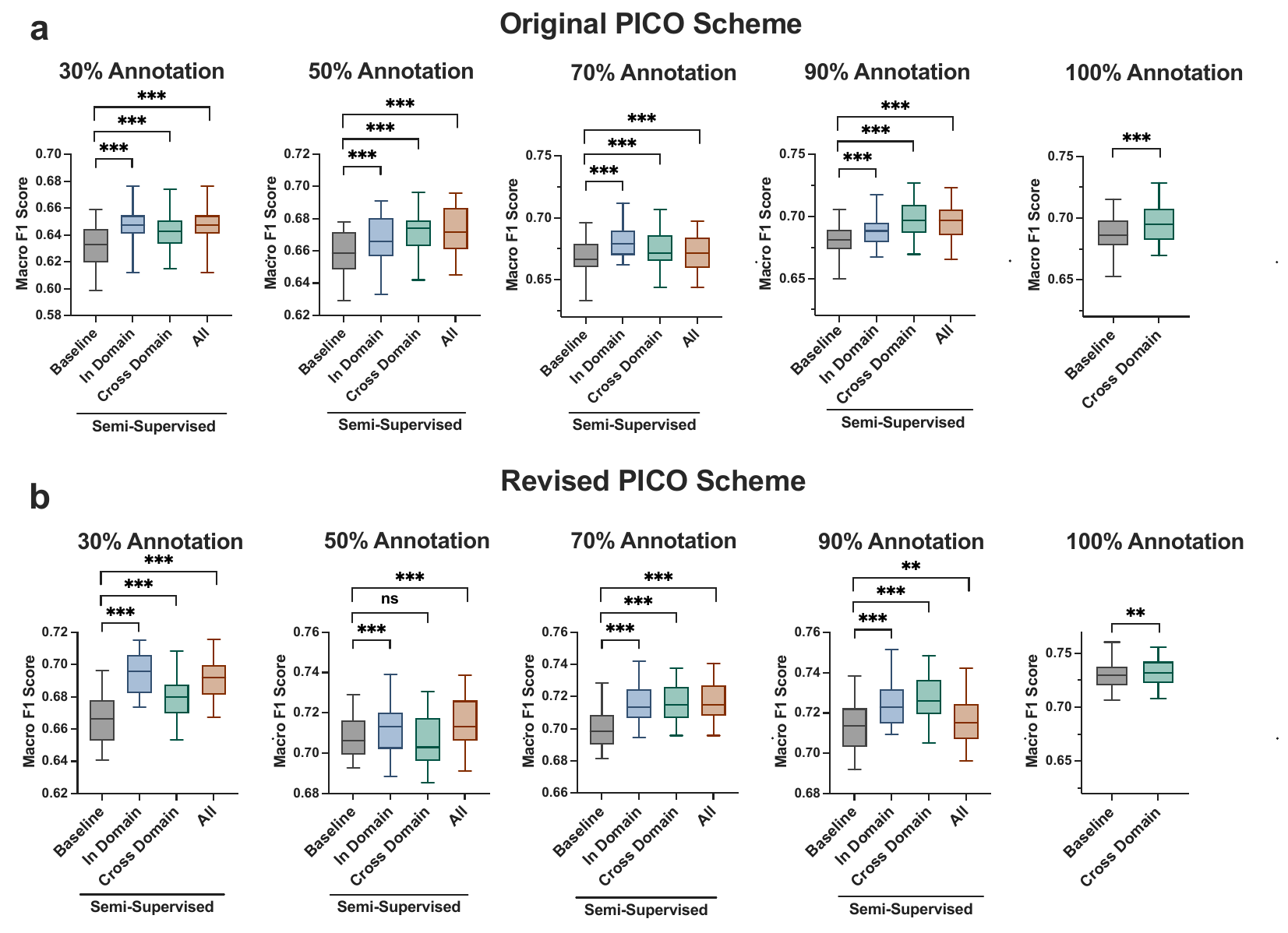}
    \caption{(a) Statistical performance comparison to baseline models in 6 simulated cases and (b) experimental setting (in-domain, cross-domain, and all) comparison. * - $p<0.05$, ** - $p<0.01$, *** - $p<0.001$.}
    \label{fig:comparison}
\end{figure}

As we increased the number of annotated samples while keeping the size of unlabeled training samples constant, we consistently observed statistically significant improvements (p-value$<$0.001) in the model's performance compared to the benchmark. These improvements were particularly notable in the extreme case when the maximum amount of labeled data was used (Figure \ref{fig:comparison}). Furthermore, the performance of the proposed algorithm consistently surpassed the baseline levels across the revised PICO scheme, showcasing the model's robustness and adaptability.

Additionally, we examined the performance differences among semi-supervised learning under various data augmentation approaches (in-domain, cross-domain, all-domain). In the original PICO scheme, models trained on both cross-domain and all-domain data performed statistically better than models trained using in-domain data (p-value \textless{} 0.001), whereas, in the revised scheme, we observed the opposite trend.

\subsubsection{Evaluation on the Independent Testing Sets}\label{evaluation-on-the-independent-testing-sets}

We further applied the best-performing model to another independent testing corpus (AD, COVID-19) \cite{hu2023towards}, and the averaged performances over 30 bootstrapped samples, along with the baseline levels, were recorded in {Table \ref{tab:Partial}. The proposed model demonstrated statistically significant improvement (p-value = 0.014 in the original scheme and p-value = 0.025 in the revised scheme) over the baselines evaluated under AD and COVID-19 corpus.

\begin{table}
\centering
\caption{Partial-matching performances of the optimal model evaluated on the external testing corpus (i.e., AD and COVID-19 datasets). The results were the average performances, and the 95\% confidence interval obtained from bootstrapped samples with 30 iterations.}
\label{tab:Partial}
\small
\begin{tabular}{lcccccc}
\toprule
Models & \multicolumn{3}{c}{Original Scheme} & \multicolumn{3}{c}{Revised Scheme} \\
\cmidrule(rl){2-4}\cmidrule(rl){5-7}
 & R & P & F1 & R & P & F1 \\
\midrule
Baseline & \makecell[t]{0.922\\(0.902, 0.943)} & \makecell[t]{0.780\\(0.756, 0.804)} & \makecell[t]{0.845\\(0.825, 0.870)} & \makecell[t]{0.931\\(0.913, 0.95)} & \makecell[t]{0.778\\(0.745, 0.811)} & \makecell[t]{0.848\\(0.825, 0.870)} \\
FinePICO & \makecell[t]{0.919\\(0.896, 0.943)} & \makecell[t]{0.795\\  (0.762, 0.830)} & \makecell[t]{0.853\\(0.826, 0.879)} & \makecell[t]{0.928\\(0.91, 0.946)} & \makecell[t]{0.789\\(0.767, 0.811)} & \makecell[t]{0.853\\(0.826, 0.879)}\\
\bottomrule
\end{tabular}
\end{table}

\subsection{Error Analysis}

We conducted an error analysis of our optimal model on 100 sentences randomly selected from the test set and identified the following error categories: 1) boundary detection error (n =14), 2) entity misclassification (n=10), and 3) failure to detect the presence of the entity (n=9). Examples of these categories are summarized in Supplementary Table \ref{sup tab:err_analysis}. Boundary detection errors are the most prevalent, suggesting that the model often failed to capture the complete entity span, especially in the names of the intervention arms and measured outcome. For instance, in the sentence ``A key secondary endpoint was the feasibility of achieving 12 meth/week (metabolic equivalent of task hours per week)'', the outcome measured was annotated as the entire phrase ``feasibility of achieving 12 meth/week (metabolic equivalent of task hours per week)''. However, our model failed to identify the content within the paratheses as part of the outcome name. Entity misclassification was the second most common error, occurring when the model incorrectly assigned values to different arms; for example, it misclassified intervention outcomes values as belonging to the control arm.

\section{Discussion}\label{discussion}

In this study, we developed a semi-supervised learning approach to overcome several key challenges in fine-grained PICO entity recognition, including the limited amount of high-quality annotated data and the lack of standardized fine-grained PICO annotation guidelines. These limitations have historically hindered the adaptability and generalizability of existing PICO extraction models.

FinePICO demonstrated substantial improvements (p-value \textless{} 0.001) compared to the baseline models across various experimental settings, including in-domain, cross-domain, and all-domain datasets. This was especially evident in scenarios where a large percentage of trained samples were unannotated. For instance, in the case where only 10\% of the training sample was labeled, FinePICO demonstrated an overall improvement of over 16\% in F1 score compared to the conventional supervised learning-based approach (in the original PICO scheme, our best model using a GPT-based label selector achieved an average F1 of 0.60 versus 0.437 for the baseline model, p-value \textless{} 0.001). FinePICO also consistently outperformed the benchmarks when applied to the revised PICO scheme, demonstrating its robustness and adaptability to varied annotation guidelines. This flexibility allows users to use their preferred fine-grained PICO scheme. As shown in the experiments (Figure \ref{fig:comparison}), the proposed algorithm effectively enhanced the model performance by augmenting training samples without needing an additional manual labeling process, significantly surpassing the models trained exclusively on fully annotated datasets.

Prior research \cite{hong2015decoupled, banitalebi2021knowledge, chen2023boosting} suggested that leveraging abundant unlabeled data with a small portion of labeled data can greatly improve learning performance. Conversely, in certain situations, semi-supervised learning offers no benefits and may even lead to performance degradation. Such situations include distribution mismatches between labeled and unlabeled data or when the labeled or unlabeled datasets are too small to extract any meaningful patterns and information \cite{oliver2018realistic, singh2008unlabeled}. Additionally, in SSL algorithms, models are iteratively retrained with newly generated pseudo-labels to enhance the prediction outcomes. However, this process can potentially introduce or amplify error propagation and class imbalance. In this study, we implemented several quality-checking strategies (i.e., a class-wise confident-based method, and label checking via large language model distillation) to minimize the noise introduced by pseudo-labels. These quality enhancement approaches are simple yet effective, achieving better performance (p-value \textless{} 0.05) than the model without such quality control measures (\textbf{Supplementary Figure \ref{supfig:annotation_example}}).

The outcomes from the study revealed the feasibility of using a semi-supervised learning-based approach to optimize fine-grained PICO entity recognition. In our experiments, we also compared the performances of the models using unlabeled datasets from three different sources: in-domain (similar domain as the labeled data), cross-domain (different domain from the labeled data), and a combination of both. In the original PICO scheme, the models trained with cross-domain data consistently exhibited better (p-value \textless0.01) performances than those trained with in-domain data. This improvement may be due to the increased data diversity and the introduction of new useful context information. These findings suggested the potential of using published cross-domain RCTs to enhance PICO extraction, especially when in-domain RCT studies were scarce.

Despite the promising results of our model, several major types of errors were recognized. The major was the boundary detection error, which is common in other PICO NER models as well \cite{hu2023towards, dhrangadhariya2023not}. Part of these errors is potentially attributed to a lack of consistency in human annotation. We believe that a clear annotation guideline that explicitly defines what to include and exclude in the labeling process can minimize these errors.

Second, for certain cases, we perceived that our model has difficulty differentiating between values in the intervention arm and control group. For example, the sentence ``Patients were randomized to receive zoledronic acid administered intravenously every 4 weeks (n = 911) vs every 12 weeks (n = 911) for 2 years'' from the RCT aims to compare the effect of a longer dosing interval (12 weeks) versus the standard dosing interval (every 4 weeks). The model misannotated the first ``911'' as the intervention sample size, and the second ``911'' as the control sample size; however, such confusion was understandable. It is also challenging for humans to make this decision without considering broader contextual information. To improve future outcomes, performing PICO recognition on a wider contextual level, rather than limiting it to the sentence level, may mitigate this confusion.

Lastly, we noticed that our model often confused with background information as one of the PICO population entities (e.g., sex, race). Such as in the sentence ``breast cancer, with an incidence of 32\%, is the most frequent cancer among Egyptian women'' which depicts the general information of breast cancer in a subpopulation, our model identified the ``Egyptian women'' as the recruited population demographic characteristics. Even though the main recruited participants were under this category, it is inaccurate to assume the study recruited participants to match the population mentioned in the background section. Thus, it is beneficial to leverage section information in determining final participants and reported results. Recently, \citet{hu2023towards2} developed a few-shot prompt learning-based approach to classifying sentences in RCTs into different subsections (Introduction, Background, Methods, Results). This demonstrates state-of-the-art performance with minimal training samples required. In the future, we can potentially incorporate the sentence classifiers before applying fine-grained PICO extractors. Another limitation of the proposed algorithm is its scalability, especially as datasets grow larger, a challenge encountered by most algorithms. The computational demands of training SSL models can strain available resources, hindering the accessibility of organizations with financial constraints and inadequate support of computational infrastructure. Future research could focus on the optimization of SSL algorithms, such as distributed training and adaptive sampling methods to prioritize more informative data during the training process.''

\section{Conclusion}\label{conclusion}

In this paper, we proposed a semi-supervised learning approach to address two notable challenges in fine-grained PICO extraction: the scarcity of high-quality annotation samples and the absence of standardized annotation guidelines. To our knowledge, this is the first attempt to comprehensively examine the performance of semi-supervised learning in fine-grained PICO extraction across various experimental settings. The findings suggested that leveraging the SSL model can effectively enhance the performance of traditional supervised learning-based models by augmenting training datasets without relying on extensive human annotation. The approach exhibited superior results compared to the benchmark, with high robustness and generalizability to other user-defined annotation schemes. This encourages the adoption of SSL techniques in extracting fine-grained PICO entities from RCTs and inspires more innovative SSL algorithms in this field.

\paragraph{Author Contributions}\label{author-contributions}

FC: Conceptualization, data curation, formal analysis, investigation, methodology, software, validation, visualization, writing-original draft, writing-review \& editing. GZ: Conceptualization, data curation, formal analysis, investigation, methodology, software, validation, visualization, writing-original draft, writing-review \& editing. YF: Data curation, investigation, validation, writing-review \& editing. YP: Conceptualization, funding acquisition, investigation, methodology, project administration, resources, supervision, validation, writing-review \& edition. CW: Conceptualization, funding acquisition, investigation, methodology, project administration, resources, supervision, validation, writing-review \& edition. All authors have read and approved the manuscript.

\paragraph{Competing Interests}\label{competing-interests}

All authors declare no competing interests.

\paragraph{Fundings}\label{fundings}

This project was sponsored by the National Library of Medicine grant R01LM014344, R01LM014573, R01LM009886, T15LM007079, the National Human Genome Research Institute grant R01HG012655, and the National Center for Advancing Translational Sciences UL1TR001873 and UL1TR002384. The content is solely the responsibility of the authors and does not necessarily represent the official views of the National Institutes of Health.

\paragraph{Data availability}\label{data-availability}

The data and codes underlying the study will be available upon request.

\bibliographystyle{unsrtnat}
\bibliography{ref}

\begin{thebibliography}{59}
\providecommand{\natexlab}[1]{#1}
\providecommand{\url}[1]{\texttt{#1}}
\expandafter\ifx\csname urlstyle\endcsname\relax
  \providecommand{\doi}[1]{doi: #1}\else
  \providecommand{\doi}{doi: \begingroup \urlstyle{rm}\Url}\fi

\bibitem[Collins(2007)]{collins2007evidence}
Jannette Collins.
\newblock Evidence-based medicine.
\newblock \emph{Journal of the American College of Radiology}, 4\penalty0 (8):\penalty0 551--554, 2007.

\bibitem[You(2016)]{you2016perspective}
Suning You.
\newblock Perspective and future of evidence-based medicine.
\newblock \emph{Stroke and vascular neurology}, 1\penalty0 (4), 2016.

\bibitem[Akobeng(2005)]{akobeng2005principles}
Anthony~K Akobeng.
\newblock Principles of evidence based medicine.
\newblock \emph{Archives of disease in childhood}, 90\penalty0 (8):\penalty0 837--840, 2005.

\bibitem[Peng et~al.(2023)Peng, Rousseau, Shortliffe, and Weng]{peng2023ai}
Yifan Peng, Justin~F Rousseau, Edward~H Shortliffe, and Chunhua Weng.
\newblock Ai-generated text may have a role in evidence-based medicine.
\newblock \emph{Nature medicine}, 29\penalty0 (7):\penalty0 1593--1594, 2023.

\bibitem[Zhang et~al.(2024{\natexlab{a}})Zhang, Jin, McInerney, Chen, Wang, Cole, Yang, Wang, Malin, Peleg, et~al.]{zhang2024leveraging}
Gongbo Zhang, Qiao Jin, Denis~Jered McInerney, Yong Chen, Fei Wang, Curtis~L Cole, Qian Yang, Yanshan Wang, Bradley~A Malin, Mor Peleg, et~al.
\newblock Leveraging generative ai for clinical evidence synthesis needs to ensure trustworthiness.
\newblock \emph{Journal of Biomedical Informatics}, 153:\penalty0 104640, 2024{\natexlab{a}}.

\bibitem[Berlin and Golub(2014)]{berlin2014meta}
Jesse~A Berlin and Robert~M Golub.
\newblock Meta-analysis as evidence: building a better pyramid.
\newblock \emph{Jama}, 312\penalty0 (6):\penalty0 603--606, 2014.

\bibitem[Cook et~al.(1997)Cook, Mulrow, and Haynes]{cook1997systematic}
Deborah~J Cook, Cynthia~D Mulrow, and R~Brian Haynes.
\newblock Systematic reviews: synthesis of best evidence for clinical decisions.
\newblock \emph{Annals of internal medicine}, 126\penalty0 (5):\penalty0 376--380, 1997.

\bibitem[Bastian et~al.(2010)Bastian, Glasziou, and Chalmers]{bastian2010seventy}
Hilda Bastian, Paul Glasziou, and Iain Chalmers.
\newblock Seventy-five trials and eleven systematic reviews a day: how will we ever keep up?
\newblock \emph{PLoS medicine}, 7\penalty0 (9):\penalty0 e1000326, 2010.

\bibitem[Borah et~al.(2017)Borah, Brown, Capers, and Kaiser]{borah2017analysis}
Rohit Borah, Andrew~W Brown, Patrice~L Capers, and Kathryn~A Kaiser.
\newblock Analysis of the time and workers needed to conduct systematic reviews of medical interventions using data from the prospero registry.
\newblock \emph{BMJ open}, 7\penalty0 (2):\penalty0 e012545, 2017.

\bibitem[Dawes et~al.(2007)Dawes, Pluye, Shea, Grad, Greenberg, and Nie]{dawes2007identification}
Martin Dawes, Pierre Pluye, Laura Shea, Roland Grad, Arlene Greenberg, and Jian-Yun Nie.
\newblock The identification of clinically important elements within medical journal abstracts: Patient--population--problem, exposure--intervention, comparison, outcome, duration and results (pecodr).
\newblock \emph{Informatics in Primary care}, 15\penalty0 (1), 2007.

\bibitem[Demner-Fushman and Lin(2007)]{demner2007answering}
Dina Demner-Fushman and Jimmy Lin.
\newblock Answering clinical questions with knowledge-based and statistical techniques.
\newblock \emph{Computational Linguistics}, 33\penalty0 (1):\penalty0 63--103, 2007.

\bibitem[Chabou and Iglewski(2018)]{chabou2018combination}
Samir Chabou and Michal Iglewski.
\newblock Combination of conditional random field with a rule based method in the extraction of pico elements.
\newblock \emph{BMC medical informatics and decision making}, 18:\penalty0 1--14, 2018.

\bibitem[Jin and Szolovits(2018)]{jin2018pico}
Di~Jin and Peter Szolovits.
\newblock Pico element detection in medical text via long short-term memory neural networks.
\newblock In \emph{Proceedings of the BioNLP 2018 workshop}, pages 67--75, 2018.

\bibitem[Graves and Schmidhuber(2005)]{graves2005framewise}
Alex Graves and J{\"u}rgen Schmidhuber.
\newblock Framewise phoneme classification with bidirectional lstm and other neural network architectures.
\newblock \emph{Neural networks}, 18\penalty0 (5-6):\penalty0 602--610, 2005.

\bibitem[Ma(2016)]{ma2016end}
X~Ma.
\newblock End-to-end sequence labeling via bi-directional lstm-cnns-crf.
\newblock \emph{arXiv preprint arXiv:1603.01354}, 2016.

\bibitem[Jin and Szolovits(2020)]{jin2020advancing}
Di~Jin and Peter Szolovits.
\newblock Advancing pico element detection in biomedical text via deep neural networks.
\newblock \emph{Bioinformatics}, 36\penalty0 (12):\penalty0 3856--3862, 2020.

\bibitem[Brockmeier et~al.(2019)Brockmeier, Ju, Przyby{\l}a, and Ananiadou]{brockmeier2019improving}
Austin~J Brockmeier, Meizhi Ju, Piotr Przyby{\l}a, and Sophia Ananiadou.
\newblock Improving reference prioritisation with pico recognition.
\newblock \emph{BMC medical informatics and decision making}, 19:\penalty0 1--14, 2019.

\bibitem[Devlin(2018)]{devlin2018bert}
Jacob Devlin.
\newblock Bert: Pre-training of deep bidirectional transformers for language understanding.
\newblock \emph{arXiv preprint arXiv:1810.04805}, 2018.

\bibitem[Lee et~al.(2020)Lee, Yoon, Kim, Kim, Kim, So, and Kang]{lee2020biobert}
Jinhyuk Lee, Wonjin Yoon, Sungdong Kim, Donghyeon Kim, Sunkyu Kim, Chan~Ho So, and Jaewoo Kang.
\newblock Biobert: a pre-trained biomedical language representation model for biomedical text mining.
\newblock \emph{Bioinformatics}, 36\penalty0 (4):\penalty0 1234--1240, 2020.

\bibitem[Liu(2019)]{liu2019roberta}
Yinhan Liu.
\newblock Roberta: A robustly optimized bert pretraining approach.
\newblock \emph{arXiv preprint arXiv:1907.11692}, 364, 2019.

\bibitem[Beltagy et~al.(2019)Beltagy, Lo, and Cohan]{beltagy2019scibert}
Iz~Beltagy, Kyle Lo, and Arman Cohan.
\newblock Scibert: A pretrained language model for scientific text.
\newblock \emph{arXiv preprint arXiv:1903.10676}, 2019.

\bibitem[Nye et~al.(2018)Nye, Li, Patel, Yang, Marshall, Nenkova, and Wallace]{nye2018corpus}
Benjamin Nye, Junyi~Jessy Li, Roma Patel, Yinfei Yang, Iain~J Marshall, Ani Nenkova, and Byron~C Wallace.
\newblock A corpus with multi-level annotations of patients, interventions and outcomes to support language processing for medical literature.
\newblock In \emph{Proceedings of the conference. Association for Computational Linguistics. Meeting}, volume 2018, page 197. NIH Public Access, 2018.

\bibitem[Hu et~al.(2023{\natexlab{a}})Hu, Keloth, Raja, Chen, and Xu]{hu2023towards}
Yan Hu, Vipina~K Keloth, Kalpana Raja, Yong Chen, and Hua Xu.
\newblock Towards precise pico extraction from abstracts of randomized controlled trials using a section-specific learning approach.
\newblock \emph{Bioinformatics}, 39\penalty0 (9):\penalty0 btad542, 2023{\natexlab{a}}.

\bibitem[Lee and Sun(2019)]{lee2019study}
Grace~E Lee and Aixin Sun.
\newblock A study on agreement in pico span annotations.
\newblock In \emph{Proceedings of the 42nd International ACM SIGIR Conference on Research and Development in Information Retrieval}, pages 1149--1152, 2019.

\bibitem[Abaho et~al.(2019)Abaho, Bollegala, Williamson, and Dodd]{abaho2019correcting}
Micheal Abaho, Danushka Bollegala, Paula~R Williamson, and Susanna Dodd.
\newblock Correcting crowdsourced annotations to improve detection of outcome types in evidence based medicine.
\newblock In \emph{KDH@ IJCAI}, pages 1--5, 2019.

\bibitem[Dhrangadhariya and M{\"u}ller(2023)]{dhrangadhariya2023not}
Anjani Dhrangadhariya and Henning M{\"u}ller.
\newblock Not so weak pico: leveraging weak supervision for participants, interventions, and outcomes recognition for systematic review automation.
\newblock \emph{JAMIA open}, 6\penalty0 (1):\penalty0 ooac107, 2023.

\bibitem[Sanchez-Graillet et~al.(2022)Sanchez-Graillet, Witte, Grimm, and Cimiano]{sanchez2022annotated}
Olivia Sanchez-Graillet, Christian Witte, Frank Grimm, and Philipp Cimiano.
\newblock An annotated corpus of clinical trial publications supporting schema-based relational information extraction.
\newblock \emph{Journal of Biomedical Semantics}, 13\penalty0 (1):\penalty0 14, 2022.

\bibitem[Ahn and Kang(2018)]{ahn2018introduction}
EunJin Ahn and Hyun Kang.
\newblock Introduction to systematic review and meta-analysis.
\newblock \emph{Korean journal of anesthesiology}, 71\penalty0 (2):\penalty0 103--112, 2018.

\bibitem[Mutinda et~al.(2022{\natexlab{a}})Mutinda, Liew, Yada, Wakamiya, and Aramaki]{mutinda2022pico}
Faith Mutinda, Kongmeng Liew, Shuntaro Yada, Shoko Wakamiya, and Eiji Aramaki.
\newblock Pico corpus: a publicly available corpus to support automatic data extraction from biomedical literature.
\newblock In \emph{Proceedings of the first Workshop on Information Extraction from Scientific Publications}, pages 26--31, 2022{\natexlab{a}}.

\bibitem[Adeva et~al.(2014)Adeva, Atxa, Carrillo, and Zengotitabengoa]{adeva2014automatic}
JJ~Garc{\'\i}a Adeva, JM~Pikatza Atxa, M~Ubeda Carrillo, and E~Ansuategi Zengotitabengoa.
\newblock Automatic text classification to support systematic reviews in medicine.
\newblock \emph{Expert Systems with Applications}, 41\penalty0 (4):\penalty0 1498--1508, 2014.

\bibitem[Yang et~al.(2022)Yang, Song, King, and Xu]{yang2022survey}
Xiangli Yang, Zixing Song, Irwin King, and Zenglin Xu.
\newblock A survey on deep semi-supervised learning.
\newblock \emph{IEEE Transactions on Knowledge and Data Engineering}, 35\penalty0 (9):\penalty0 8934--8954, 2022.

\bibitem[Yang et~al.(2023)Yang, Zhang, Chen, Wang, and Kim]{yang2023prototype}
Weiyi Yang, Richong Zhang, Junfan Chen, Lihong Wang, and Jaein Kim.
\newblock Prototype-guided pseudo labeling for semi-supervised text classification.
\newblock In \emph{Proceedings of the 61st Annual Meeting of the Association for Computational Linguistics (Volume 1: Long Papers)}, pages 16369--16382, 2023.

\bibitem[Hong et~al.(2015)Hong, Noh, and Han]{hong2015decoupled}
Seunghoon Hong, Hyeonwoo Noh, and Bohyung Han.
\newblock Decoupled deep neural network for semi-supervised semantic segmentation.
\newblock \emph{Advances in neural information processing systems}, 28, 2015.

\bibitem[Roli and Marcialis(2006)]{roli2006semi}
Fabio Roli and Gian~Luca Marcialis.
\newblock Semi-supervised pca-based face recognition using self-training.
\newblock In \emph{Structural, Syntactic, and Statistical Pattern Recognition: Joint IAPR International Workshops, SSPR 2006 and SPR 2006, Hong Kong, China, August 17-19, 2006. Proceedings}, pages 560--568. Springer, 2006.

\bibitem[Bickel(2006)]{bickel2006ecml}
Steffen Bickel.
\newblock Ecml-pkdd discovery challenge 2006 overview.
\newblock In \emph{ECML-PKDD Discovery Challenge Workshop}, pages 1--9, 2006.

\bibitem[Li et~al.(2009)Li, Li, and Zhou]{li2009semi}
Ming Li, Hang Li, and Zhi-Hua Zhou.
\newblock Semi-supervised document retrieval.
\newblock \emph{Information Processing \& Management}, 45\penalty0 (3):\penalty0 341--355, 2009.

\bibitem[Duh and Kirchhoff(2011)]{duh2011semi}
Kevin Duh and Katrin Kirchhoff.
\newblock Semi-supervised ranking for document retrieval.
\newblock \emph{Computer Speech \& Language}, 25\penalty0 (2):\penalty0 261--281, 2011.

\bibitem[Erkan et~al.(2007)Erkan, {\"O}zg{\"u}r, and Radev]{erkan2007semi}
Gunes Erkan, Arzucan {\"O}zg{\"u}r, and Dragomir Radev.
\newblock Semi-supervised classification for extracting protein interaction sentences using dependency parsing.
\newblock In \emph{Proceedings of the 2007 Joint Conference on Empirical Methods in Natural Language Processing and Computational Natural Language Learning (EMNLP-CoNLL)}, pages 228--237, 2007.

\bibitem[Gu et~al.(2021)Gu, Tinn, Cheng, Lucas, Usuyama, Liu, Naumann, Gao, and Poon]{gu2021domain}
Yu~Gu, Robert Tinn, Hao Cheng, Michael Lucas, Naoto Usuyama, Xiaodong Liu, Tristan Naumann, Jianfeng Gao, and Hoifung Poon.
\newblock Domain-specific language model pretraining for biomedical natural language processing.
\newblock \emph{ACM Transactions on Computing for Healthcare (HEALTH)}, 3\penalty0 (1):\penalty0 1--23, 2021.

\bibitem[Ferreira et~al.(2023)Ferreira, Lee, and D{\'o}rea]{ferreira2023using}
Rafael~EP Ferreira, Yong~Jae Lee, and Jo{\~a}o~RR D{\'o}rea.
\newblock Using pseudo-labeling to improve performance of deep neural networks for animal identification.
\newblock \emph{Scientific Reports}, 13\penalty0 (1):\penalty0 13875, 2023.

\bibitem[Xie et~al.(2020)Xie, Dai, Hovy, Luong, and Le]{xie2020unsupervised}
Qizhe Xie, Zihang Dai, Eduard Hovy, Thang Luong, and Quoc Le.
\newblock Unsupervised data augmentation for consistency training.
\newblock \emph{Advances in neural information processing systems}, 33:\penalty0 6256--6268, 2020.

\bibitem[Zhang et~al.(2021)Zhang, Lin, Han, and Sun]{zhang2021biasing}
Wenkai Zhang, Hongyu Lin, Xianpei Han, and Le~Sun.
\newblock De-biasing distantly supervised named entity recognition via causal intervention.
\newblock \emph{arXiv preprint arXiv:2106.09233}, 2021.

\bibitem[Hu et~al.(2024)Hu, Chen, Du, Peng, Keloth, Zuo, Zhou, Li, Jiang, Lu, et~al.]{hu2024improving}
Yan Hu, Qingyu Chen, Jingcheng Du, Xueqing Peng, Vipina~Kuttichi Keloth, Xu~Zuo, Yujia Zhou, Zehan Li, Xiaoqian Jiang, Zhiyong Lu, et~al.
\newblock Improving large language models for clinical named entity recognition via prompt engineering.
\newblock \emph{Journal of the American Medical Informatics Association}, page ocad259, 2024.

\bibitem[Zhang et~al.(2024{\natexlab{b}})Zhang, Zhou, Hu, Xu, Weng, and Peng]{zhang2024span}
Gongbo Zhang, Yiliang Zhou, Yan Hu, Hua Xu, Chunhua Weng, and Yifan Peng.
\newblock A span-based model for extracting overlapping pico entities from randomized controlled trial publications.
\newblock \emph{Journal of the American Medical Informatics Association}, 31\penalty0 (5):\penalty0 1163--1171, 2024{\natexlab{b}}.

\bibitem[Mutinda et~al.(2022{\natexlab{b}})Mutinda, Liew, Yada, Wakamiya, and Aramaki]{mutinda2022automatic}
Faith~Wavinya Mutinda, Kongmeng Liew, Shuntaro Yada, Shoko Wakamiya, and Eiji Aramaki.
\newblock Automatic data extraction to support meta-analysis statistical analysis: a case study on breast cancer.
\newblock \emph{BMC Medical Informatics and Decision Making}, 22\penalty0 (1):\penalty0 158, 2022{\natexlab{b}}.

\bibitem[Bird et~al.(2009)Bird, Klein, and Loper]{bird2009natural}
Steven Bird, Ewan Klein, and Edward Loper.
\newblock \emph{Natural language processing with Python: analyzing text with the natural language toolkit}.
\newblock " O'Reilly Media, Inc.", 2009.

\bibitem[Sang and Buchholz(2000)]{sang2000introduction}
Erik~F Sang and Sabine Buchholz.
\newblock Introduction to the conll-2000 shared task: Chunking.
\newblock \emph{arXiv preprint cs/0009008}, 2000.

\bibitem[He et~al.(2023)He, Wang, Lu, Lin, Han, Sun, and Sun]{he2023document}
Shaojie He, Tianshu Wang, Yaojie Lu, Hongyu Lin, Xianpei Han, Yingfei Sun, and Le~Sun.
\newblock Document information extraction via global tagging.
\newblock In \emph{China National Conference on Chinese Computational Linguistics}, pages 145--158. Springer, 2023.

\bibitem[Wang et~al.(2023)Wang, Liu, Ying, Yang, Chen, Liu, Zhang, Yan, Lu, Gao, et~al.]{wang2023optimized}
Guangyu Wang, Xiaohong Liu, Zhen Ying, Guoxing Yang, Zhiwei Chen, Zhiwen Liu, Min Zhang, Hongmei Yan, Yuxing Lu, Yuanxu Gao, et~al.
\newblock Optimized glycemic control of type 2 diabetes with reinforcement learning: a proof-of-concept trial.
\newblock \emph{Nature Medicine}, 29\penalty0 (10):\penalty0 2633--2642, 2023.

\bibitem[Assmann et~al.(2000)Assmann, Pocock, Enos, and Kasten]{assmann2000subgroup}
Susan~F Assmann, Stuart~J Pocock, Laura~E Enos, and Linda~E Kasten.
\newblock Subgroup analysis and other (mis) uses of baseline data in clinical trials.
\newblock \emph{The Lancet}, 355\penalty0 (9209):\penalty0 1064--1069, 2000.

\bibitem[Bhandari et~al.(2006)Bhandari, Devereaux, Li, Mah, Lim, Sch{\"u}nemann, and Tornetta~III]{bhandari2006misuse}
Mohit Bhandari, PJ~Devereaux, Patricia Li, Doug Mah, Ki~Lim, Holger~J Sch{\"u}nemann, and Paul Tornetta~III.
\newblock Misuse of baseline comparison tests and subgroup analyses in surgical trials.
\newblock \emph{Clinical Orthopaedics and Related Research{\textregistered}}, 447:\penalty0 247--251, 2006.

\bibitem[Nakayama(2018)]{nakayama2018seqeval}
Hiroki Nakayama.
\newblock seqeval: A python framework for sequence labeling evaluation.
\newblock \emph{Software available from https://github. com/chakki-works/seqeval}, 2018.

\bibitem[Heddes et~al.(2021)Heddes, Meerdink, Pieters, and Marx]{heddes2021automatic}
Jenny Heddes, Pim Meerdink, Miguel Pieters, and Maarten Marx.
\newblock The automatic detection of dataset names in scientific articles.
\newblock \emph{Data}, 6\penalty0 (8):\penalty0 84, 2021.

\bibitem[Seki and Mostafa(2003)]{seki2003probabilistic}
Kazuhiro Seki and Javed Mostafa.
\newblock A probabilistic model for identifying protein names and their name boundaries.
\newblock In \emph{Computational Systems Bioinformatics. CSB2003. Proceedings of the 2003 IEEE Bioinformatics Conference. CSB2003}, pages 251--258. IEEE, 2003.

\bibitem[Banitalebi-Dehkordi(2021)]{banitalebi2021knowledge}
Amin Banitalebi-Dehkordi.
\newblock Knowledge distillation for low-power object detection: A simple technique and its extensions for training compact models using unlabeled data.
\newblock In \emph{Proceedings of the IEEE/CVF International Conference on Computer Vision}, pages 769--778, 2021.

\bibitem[Chen et~al.(2023)Chen, Tan, Zhao, Chen, Song, Liang, and Lu]{chen2023boosting}
Yuhao Chen, Xin Tan, Borui Zhao, Zhaowei Chen, Renjie Song, Jiajun Liang, and Xuequan Lu.
\newblock Boosting semi-supervised learning by exploiting all unlabeled data.
\newblock In \emph{Proceedings of the IEEE/CVF Conference on Computer Vision and Pattern Recognition}, pages 7548--7557, 2023.

\bibitem[Oliver et~al.(2018)Oliver, Odena, Raffel, Cubuk, and Goodfellow]{oliver2018realistic}
Avital Oliver, Augustus Odena, Colin~A Raffel, Ekin~Dogus Cubuk, and Ian Goodfellow.
\newblock Realistic evaluation of deep semi-supervised learning algorithms.
\newblock \emph{Advances in neural information processing systems}, 31, 2018.

\bibitem[Singh et~al.(2008)Singh, Nowak, and Zhu]{singh2008unlabeled}
Aarti Singh, Robert Nowak, and Jerry Zhu.
\newblock Unlabeled data: Now it helps, now it doesn't.
\newblock \emph{Advances in neural information processing systems}, 21, 2008.

\bibitem[Hu et~al.(2023{\natexlab{b}})Hu, Chen, and Xu]{hu2023towards2}
Yan Hu, Yong Chen, and Hua Xu.
\newblock Towards more generalizable and accurate sentence classification in medical abstracts with less data.
\newblock \emph{Journal of Healthcare Informatics Research}, 7\penalty0 (4):\penalty0 542--556, 2023{\natexlab{b}}.

\end{thebibliography}

\newpage
\appendix
\setcounter{table}{0}
\setcounter{figure}{0}
\renewcommand\figurename{Supplementary Figure} 
\renewcommand\tablename{Supplementary Table}

\section*{Supplementary materials}
\label{sec:appendix}

\captionsetup{type=table}
{\captionof{table}{Customized Prompts for each entity.}
\label{sup tab:prompt}
\small\rowcolors{2}{}{gray!10!white}

\begin{xltabular}{\textwidth}{>{\hsize=.5\hsize}X>{\hsize=1.5\hsize}X}
\toprule
Entity & Prompt\\
\midrule

\endfirsthead

\rowcolor{white}
\multicolumn{2}{l}%
{\tablename\ \thetable{} -- continued from previous page} \\
\toprule
\rowcolor{white}
Entity & Prompt\\
\midrule

\endhead

\bottomrule
\multicolumn{2}{r}{{Continued on next page}} \\
\endfoot

\bottomrule
\endlastfoot

Condition & Based on the entity definition below, check if the detected tokens '{tokens}' describe part of the condition in this sentence: "{sentence}". Return yes or no only.
\newline\newline
Definition: condition refers to medical conditions that patients often experience, which can be the symptoms that an RCT attempts to prevent or alleviate
\newline\newline
Sample output: Yes
\\
Eligibility & Based on the entity definition below,  check if the detected tokens '{tokens}'  describe part of  the eligibility in this sentence: "{sentence}". Return yes or no only.
\newline\newline
Definition: eligibility specifies the particular health conditions or stages of a disease, or medical history that participants must have, or medication treatment participants receive. Sex or age is not included in this category.
\newline\newline
Sample output: Yes
\\
Total sample size & Check if the detected tokens '{tokens}'  describe the total sample size of the recruited participants in this sentence: "{sentence}". Return yes or no only.
\newline\newline
Sample output: Yes
\\
Age & Check if the detected tokens '{tokens}' describe the age in this sentence: "{sentence}". Return yes or no only.
\newline\newline
Sample output: Yes
\\
Location & Check if the detected tokens '{tokens}' describe part of the location in this sentence: "{sentence}". Return yes or no only.
\newline\newline
Sample output: Yes
\\
Ethnicity & Check if the detected tokens '{tokens}' describe the ethnicity in this sentence: "{sentence}". Return yes or no only.
\newline\newline
Sample output: Yes
\\
Intervention name & Check if the detected tokens '{tokens}'  describe part of the intervention under the PICO framework in this sentence: "{sentence}". Return yes or no only.
\newline\newline
Sample output: Yes
\\
Intervention arm sample size & Check if the detected tokens '{tokens}'  describe the sample size of the intervention arm in this sentence: "{sentence}". Return yes or no only.
\newline\newline
Sample size in control arm (e.g., placebo group) should not be included. 
\newline\newline
Sample output: Yes
\\
Control name & Check if the detected tokens '{tokens}'  describe part of the control under the PICO framework  in this sentence: "{sentence}". Return yes or no only.
\newline\newline
Sample output: Yes
\\
Control arm sample size & Check if the detected tokens '{tokens}'  describe the sample size of the control arm in this sentence: "{sentence}". Return yes or no only.
\newline\newline
Sample size in intervention arm should not be included. 
\newline\newline
Sample output: Yes
\\
Outcome & Check if the detected tokens '{tokens}'  describe part of the outcome in this sentence: "{sentence}". Return yes or no only.
\newline\newline
Sample output: Yes
\\
Outcome measure & Based on the entity definition below, Check if the detected tokens '{tokens}'  describe part of the outcome measure in this sentence: "{sentence}". Return yes or no only.
\newline\newline
Definition: outcome measure refers to the metrics used to quantify the outcomes of an RCT study
\newline\newline
Sample output: Yes
\\
Discrete outcome values in intervention arm & Based on the example below, Check if the detected tokens '{tokens}'  describe the results in the intervention arm this sentence: "{sentence}". Return yes or no only.\newline\newline
Example 1: \newline
    Input:\newline
    - Check tokens: 79\newline
    - Sentence: 79 deaths were observed in the HDCT arm and 77 deaths were observed in the placebo arm.\newline
    Output: Yes\newline
\newline\newline
Example 2:\newline
    Input:\newline
    - Check tokens: 77\newline
    - Sentence: 79 deaths were observed in the HDCT arm and 77 deaths were observed in the control arm.\newline
    Output: No
\\
Discrete outcome values in control arm & Based on the example below, Check if the detected tokens '{tokens}'  describe the results in the control arm this sentence: "{sentence}". Return yes or no only.
\newline\newline
Example 1: \newline
    Input:\newline
    - Check tokens: 77\newline
    - Sentence: 79 deaths were observed in the HDCT arm and 77 deaths were observed in the ST arm\newline
    Output: Yes
\newline\newline
Example 2: \newline
    Input:\newline
    - Check tokens: 79\newline
    - Sentence: 79 deaths were observed in the HDCT arm and 77 deaths were observed in the ST arm\newline
    Output: No
\\
Continuous numeric outcome values in control arm & Check if the detected tokens '{tokens}'  describe the continuous numeric values in the control arm this sentence: "{sentence}". Return yes or no only.
\newline\newline
Sample output: Yes
\\
Continuous numeric outcome values in intervention arm & Check if the detected tokens '{tokens}'  describe the continuous numeric values in the intervention arm this sentence: "{sentence}". Return yes or no only.
\newline\newline
Sample output: Yes
\\
Standard deviation in intervention arm & Check if the detected tokens '{tokens}'  describe the standard deviation values in the intervention arm this sentence: "{sentence}". Return yes or no only.
\newline\newline
Sample output: Yes
\\
Standard deviation in control arm & Check if the detected tokens '{tokens}'  describe the standard deviation values in the control arm this sentence: "{sentence}". Return yes or no only.
\newline\newline
Sample output: Yes\\
\end{xltabular}}

\newpage

\captionof{table}{Macro-average performance of models (BioBERT, SciBERT, ClinicalBERT, BiomedBERT) on the testing set.}
\label{sup tab:performance}
{\begin{tabular}{lcccccc}
\toprule
Models & \multicolumn{3}{c}{Original Scheme} & \multicolumn{3}{c}{Revised Scheme} \\
\cmidrule(rl){2-4}\cmidrule(rl){5-7}
 & R & P & F1 & R & P & F1 \\
\midrule
Baseline & 0.922 & 0.780 & 0.845 & 0.931 & 0.778 & 0.848\\
Proposed Model & 0.919 & 0.795 & 0.853 & 0.928 & 0.789 & 0.853\\
\bottomrule
\end{tabular}}

\newpage

{\captionof{table}{Number of sentences in train, validation, and test sets in different simulated cases. }
\label{sup tab:scheme}
\small

\begin{tabular}{lrrrrrr}
\toprule
& \textbf{Case 1} & \textbf{Case 2} & \textbf{Case 3} & \textbf{Case 4} & \textbf{Case 5} & \textbf{Case 6} \\
\textbf{Category} & \textbf{10\%} & \textbf{30\%} & \textbf{50\%} & \textbf{70\%} & \textbf{90\%} & \textbf{100\%} \\
\midrule
\textbf{Train} & & & & & & \\
\textit{In Domain setting: PICO-Corpus} \\
\hspace{2em}w/ labels & 981 & 2,945 & 4,909 & 6,873 & 8,837 & - \\
\hspace{2em}w/o labels & 8,838 & 6,874 & 4,910 & 2,946 & 982 & - \\
\textit{Cross Domain setting: EBM-NLP} \\
\hspace{2em}w/ labels & 981 & 2,945 & 4,909 & 6,873 & 8,837 & 9,819 \\
\hspace{2em}w/o labels & 12,700 & 12,700 & 12,700 & 12,700 & 12,700 & 12,700 \\
\textit{All Domains setting: In Domain + Cross Domain} \\
\hspace{2em}w/ labels & 981 & 2,945 & 4,909 & 6,873 & 982 & 9,819 \\
\hspace{2em}w/o labels & 21,538 & 19,547 & 17,610 & 15,646 & 21,537 & 12,700 \\
\midrule
\textbf{Validation} & 1,091 & 1,091 & 1,091 & 1,091 & 1,091 & 1,091 \\
\midrule
\textbf{Testing corpus 1:} PICO-Corpus & 2,717 & 2,717 & 2,717 & 2,717 & 2,717 & 2,717 \\
\midrule
\textbf{Testing corpus 2:} AD + COVID-19 & 1,627 & 1,627 & 1,627 & 1,627 & 1,627 & 1,627 \\
\bottomrule
\end{tabular}%
}

\newpage

{\captionof{table}{Entity counts of modified PICO scheme.}
\label{sup tab:entity_count}
\begin{tabular}{lrrrr}
\toprule
& \textbf{PICO-Corpus} & \textbf{EBM-NLP} & \textbf{AD} & \textbf{COVID-19} \\
\midrule
\textbf{Abstracts} & \textbf{1,011} & \textbf{1,200} & \textbf{150} & \textbf{150} \\
Training & 1,010 & & & \\
Validation & 645 & & & \\
Test & 944 & & & \\
\midrule
\textbf{Population (P)} & & 3,951 & 215 & 262 \\
Total sample size & 1,094 & - & - & - \\
Sample size in INT & 887 & & & \\
Sample size in CTL & 784 & - & - & - \\
Sex & 1,991 & & & \\
Age & 231 & - & - & - \\
Eligibility condition \& criteria & 1,252 & - & - & - \\
Other demographics (location, ethnicity, etc.) & 287 & - & - & - \\
\midrule
\textbf{Intervention (I)} & 1,067 & 5,916 & 467 & 602 \\
\midrule
\textbf{Control (C)} & 979 & 563 & 103 & 180 \\
\midrule
\textbf{Outcome (O)} & & 7,151 & 626 & 626 \\
Names of study outcomes & 6,134 & - & - & - \\
Binary outcomes & & & & \\
- Absolute value, INT/CTL & 556/465 & - & - & - \\
- Percentage values, INT/CTL & 1,376/1,148 & - & - & - \\
Continuous outcomes & & & & \\
- Mean, INT/CTL & 336/327 & - & - & - \\
- Median, INT/CTL & 270/247 & - & - & - \\
- Standard deviation, INT/CTL & 129/124 & - & - & - \\
- Others, INT/CTL & 8/8 & - & - & - \\
\bottomrule
\end{tabular}
}

\newpage

{\captionof{table}{Performances of self-cleaning methods.}
\label{sup tab:clean_perf}
\footnotesize
\begin{tabular}{
c@{~~(}c@{,~}c@{)\hspace*{10pt}}
c@{~~(}c@{,~}c@{)\hspace*{10pt}}
c@{~~(}c@{,~}c@{)\hspace*{10pt}}
ccc}
\toprule
\multicolumn{3}{c}{\textbf{Recall}} & \multicolumn{3}{c}{\textbf{Precision}} &
\multicolumn{3}{c}{\textbf{F1}} & \textbf{Model type} & 
\textbf{Data augmentation} & 
\makecell[b]{\textbf{PICO}\\\textbf{scheme}}\\
\midrule
0.636 & 0.605 & 0.667 & 0.561 & 0.609 & 0.664 & 0.596 & 0.609 & 0.664 & Class adaptive & In\_Domain & raw \\
0.617 & 0.585 & 0.648 & 0.571 & 0.589 & 0.645 & 0.593 & 0.589 & 0.645 & Class adaptive & Cross\_Domain\_EBM & raw \\
0.682 & 0.654 & 0.709 & 0.626 & 0.657 & 0.707 & 0.653 & 0.657 & 0.706 & Class adaptive & In\_Domain & new \\
0.677 & 0.650 & 0.704 & 0.627 & 0.655 & 0.699 & 0.651 & 0.654 & 0.700 & Class adaptive & Cross\_Domain\_EBM & new \\
0.607 & 0.576 & 0.638 & 0.566 & 0.578 & 0.636 & 0.586 & 0.578 & 0.636 & Confident-based & In\_Domain & raw \\
0.619 & 0.590 & 0.647 & 0.580 & 0.595 & 0.643 & 0.598 & 0.594 & 0.643 & Confident-based & Cross\_Domain\_EBM & raw \\
0.675 & 0.646 & 0.705 & 0.628 & 0.648 & 0.703 & 0.651 & 0.648 & 0.703 & Confident-based & In\_Domain & new \\
0.652 & 0.626 & 0.677 & 0.613 & 0.627 & 0.676 & 0.632 & 0.628 & 0.675 & Confident-based & Cross\_Domain\_EBM & new \\
0.607 & 0.578 & 0.635 & 0.591 & 0.578 & 0.635 & 0.599 & 0.580 & 0.633 & GPT & In\_Domain & raw \\
0.636 & 0.606 & 0.667 & 0.567 & 0.610 & 0.663 & 0.600 & 0.609 & 0.664 & GPT & Cross\_Domain\_EBM & raw \\
0.639 & 0.614 & 0.664 & 0.607 & 0.612 & 0.665 & 0.622 & 0.615 & 0.663 & GPT & In\_Domain & new \\
0.613 & 0.587 & 0.638 & 0.608 & 0.586 & 0.639 & 0.610 & 0.589 & 0.637 & GPT & Cross\_Domain\_EBM & new \\
\bottomrule
\end{tabular}
}

\newpage

{\captionof{table}{Baseline model and FinePICO performance comparison.}
\label{sup tab:baseline_comp}
\footnotesize
\begin{tabular}{c*{11}{c}}
\toprule
\multirow{2}{*}{\makecell[b]{\textbf{Percent}\\\textbf{Anno}}} & \multicolumn{3}{c}{\textbf{p-value}} & \multicolumn{3}{c}{\textbf{Basline Avg}} & \multicolumn{3}{c}{\textbf{Semi Avg}}\\
\cmidrule(rl){2-4}\cmidrule(rl){5-7}\cmidrule(rl){8-10}
& \textbf{R} & \textbf{P} & \textbf{F1} & \textbf{R} & \textbf{P} & \textbf{F1} &
\textbf{R} & \textbf{P} & \textbf{F1} & \textbf{Model Type} & \textbf{Ontology} \\
\midrule
10 & 9.52E-32 & 2.82E-35 & 2.82E-34 & 0.49 & 0.39 & 0.44 & 0.61 & 0.57 & 0.59 & In\_Domain & raw \\
30 & 0.32 & 6.26E-15 & 5.84E-09 & 0.67 & 0.60 & 0.63 & 0.67 & 0.62 & 0.65 & In\_Domain & raw \\
50 & 0.97 & 1.02E-10 & 8.73E-06 & 0.69 & 0.63 & 0.66 & 0.69 & 0.65 & 0.67 & In\_Domain & raw \\
70 & 1.88E-07 & 1.15E-19 & 4.93E-16 & 0.69 & 0.63 & 0.66 & 0.70 & 0.66 & 0.68 & In\_Domain & raw \\
90 & 1E-3 & 2.96E-08 & 5.58E-07 & 0.71 & 0.65 & 0.68 & 0.72 & 0.66 & 0.69 & In\_Domain & raw \\

10 & 1.04E-36 & 1.85E-40 & 2.01E-39 & 0.49 & 0.39 & 0.44 & 0.62 & 0.58 & 0.60 & Cross\_Domain\_EBM & raw \\
30 & 0.05 & 3.57E-12 & 1.75E-08 & 0.67 & 0.60 & 0.63 & 0.67 & 0.62 & 0.64 & Cross\_Domain\_EBM & raw \\
50 & 1.50E-08 & 1.05E-12 & 7.79E-12 & 0.69 & 0.63 & 0.66 & 0.70 & 0.65 & 0.67 & Cross\_Domain\_EBM & raw \\
70 & 9.77E-10 & 9.42E-13 & 1.79E-12 & 0.69 & 0.63 & 0.66 & 0.70 & 0.65 & 0.67 & Cross\_Domain\_EBM & raw \\
90 & 2.54E-12 & 2.11E-12 & 4.98E-13 & 0.71 & 0.65 & 0.68 & 0.73 & 0.67 & 0.70 & Cross\_Domain\_EBM & raw \\
100 & 0.64 & 2.01E-08 & 1.48E-05 & 0.72 & 0.66 & 0.69 & 0.72 & 0.68 & 0.70 & Cross\_Domain\_EBM & raw \\

10 & 3.34E-32 & 3.72E-37 & 2.02E-35 & 0.49 & 0.39 & 0.44 & 0.60 & 0.56 & 0.58 & Whole & raw \\
30 & 3.74E-10 & 1.58E-20 & 1.50E-17 & 0.67 & 0.60 & 0.63 & 0.69 & 0.64 & 0.67 & Whole & raw \\
50 & 1.05E-06 & 2.53E-12 & 1.37E-10 & 0.69 & 0.63 & 0.66 & 0.70 & 0.65 & 0.67 & Whole & raw \\
70 & 9.15E-09 & 2.15E-10 & 3.30E-10 & 0.69 & 0.63 & 0.66 & 0.70 & 0.65 & 0.67 & Whole & raw \\
90 & 1E-3 & 3.83E-16 & 1.17E-12 & 0.71 & 0.65 & 0.68 & 0.72 & 0.68 & 0.70 & Whole & raw \\

10 & 1.48E-30 & 2.07E-32 & 3.06E-32 & 0.57 & 0.48 & 0.52 & 0.68 & 0.63 & 0.65 & In\_Domain & new \\
30 & 8.57E-14 & 4.26E-18 & 2.00E-17 & 0.70 & 0.63 & 0.67 & 0.72 & 0.67 & 0.69 & In\_Domain & new \\
50 & 1.19E-05 & 0.17     & 1E-03    & 0.74 & 0.68 & 0.71 & 0.75 & 0.68 & 0.71 & In\_Domain & new \\
70 & 5.82E-08 & 4.33E-18 & 5.12E-16 & 0.73 & 0.67 & 0.70 & 0.73 & 0.70 & 0.72 & In\_Domain & new \\
90 & 4.77E-07 & 1.91E-17 & 7.01E-15 & 0.74 & 0.69 & 0.71 & 0.75 & 0.70 & 0.73 & In\_Domain & new \\

10 & 1.76E-32 & 1.27E-35 & 3.49E-35 & 0.57 & 0.48 & 0.52 & 0.65 & 0.61 & 0.63 & Cross\_Domain\_EBM & new \\
30 & 2.99E-10 & 1.07E-12 & 1.76E-13 & 0.70 & 0.63 & 0.67 & 0.71 & 0.65 & 0.68 & Cross\_Domain\_EBM & new \\
50 & 1E-03    & 2.60E-10 & 0.84     & 0.74 & 0.68 & 0.71 & 0.72 & 0.69 & 0.71 & Cross\_Domain\_EBM & new \\
70 & 6.03E-13 & 4.05E-18 & 4.30E-18 & 0.73 & 0.67 & 0.70 & 0.74 & 0.70 & 0.72 & Cross\_Domain\_EBM & new \\
90 & 1.86E-08 & 1.26E-15 & 1.33E-14 & 0.74 & 0.69 & 0.71 & 0.75 & 0.71 & 0.73 & Cross\_Domain\_EBM & new \\
100 & 0.71    & 0.002    & 0.01     & 0.75 & 0.71 & 0.73 & 0.75 & 0.71 & 0.73 & Cross\_Domain\_EBM & raw \\

10 & 6.27E-32 & 3.72E-35 & 3.88E-34 & 0.57 & 0.48 & 0.52 & 0.66 & 0.61 & 0.64 & Whole & new \\
30 & 9.66E-06 & 6.70E-21 & 6.23E-17 & 0.70 & 0.63 & 0.67 & 0.71 & 0.67 & 0.69 & Whole & new \\
50 & 0.16     & 1.44E-10 & 4.66E-07 & 0.74 & 0.68 & 0.71 & 0.74 & 0.69 & 0.72 & Whole & new \\
70 & 1.82E-06 & 2.21E-16 & 2.40E-13 & 0.73 & 0.67 & 0.70 & 0.74 & 0.70 & 0.72 & Whole & new \\
90 & 0.57     & 4.56E-05 & 0.01     & 0.74 & 0.69 & 0.71 & 0.74 & 0.69 & 0.72 & Whole & new \\
\bottomrule
\end{tabular}%
}

\newpage

{\captionof{table}{Error analysis of 100 sentences randomly selected from test set. INT refers to the intervention arm. The text highlighted in red refers to instances where the model either failed to identify or incorrectly assigned it to the wrong category. The text highlighted in green represents entities that the model correctly identified and categorized.}
\label{sup tab:err_analysis}
\small
\begin{tabularx}{\textwidth}{p{10em}
>{\raggedright\arraybackslash}Xr}
\toprule
\textbf{Error Categories} & \textbf{Examples} & \textbf{Frequency} \\
\midrule
Boundary detection error & A key secondary endpoint was the \colorbox{green!30}{feasibility of achieving meth/week} (\colorbox{red!20}{metabolic equivalent of task hours per week}). & 14 \\
\midrule
Entity misclassification & We measure lipids in an average of 4.2 blood samples for \colorbox{red!20}{279} \textcolor{red}{(INT\_sample)} invasive breast cancer case subjects and \colorbox{red!20}{558} \textcolor{red}{(Control\_sample)} matched control subjects. & 10 \\
\midrule
Fails to detect the presence of the entity & Findings suggest that a pragmatic exercise intervention yields more \colorbox{green!30}{women} with markedly improved \colorbox{red!20}{quality of life} after breast cancer than usual care and may be cost-effective. & 9 \\
\bottomrule
\end{tabularx}}

\newpage

{\centering
\includegraphics[width=0.3\linewidth]{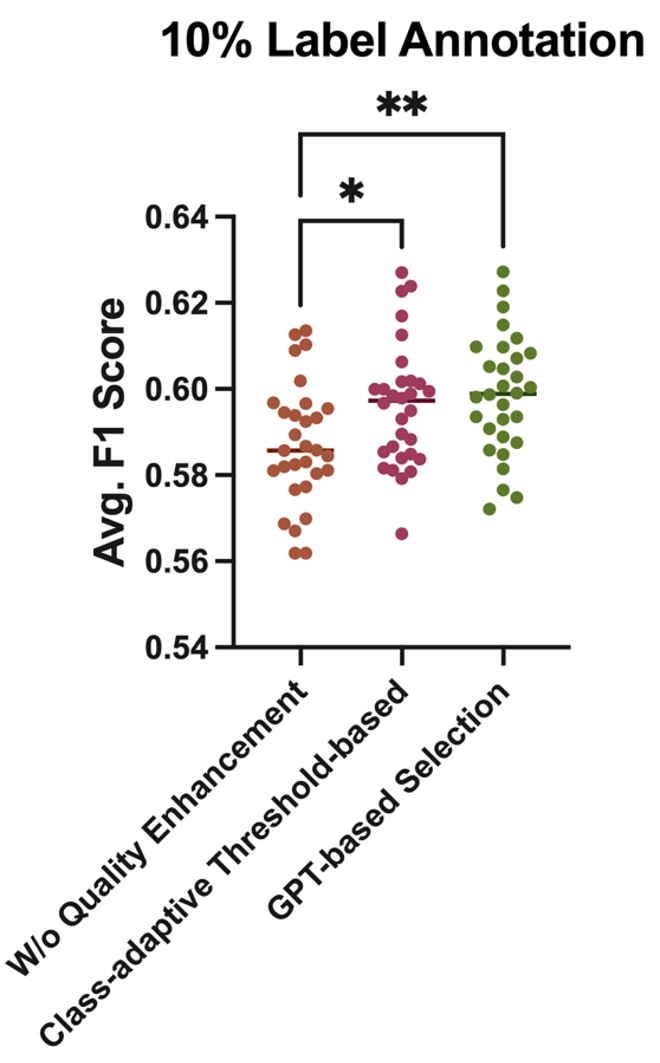}
\captionof{figure}{Assessment of quality control mechanisms. The models with quality control (threshold-based and model distillation) perform statistically better than the one without. The models were evaluated in the low-resource setting where only 10\% of sentences were labeled.}
\label{supfig:label_annotation}}

\newpage

{\centering
\includegraphics[width=\linewidth]{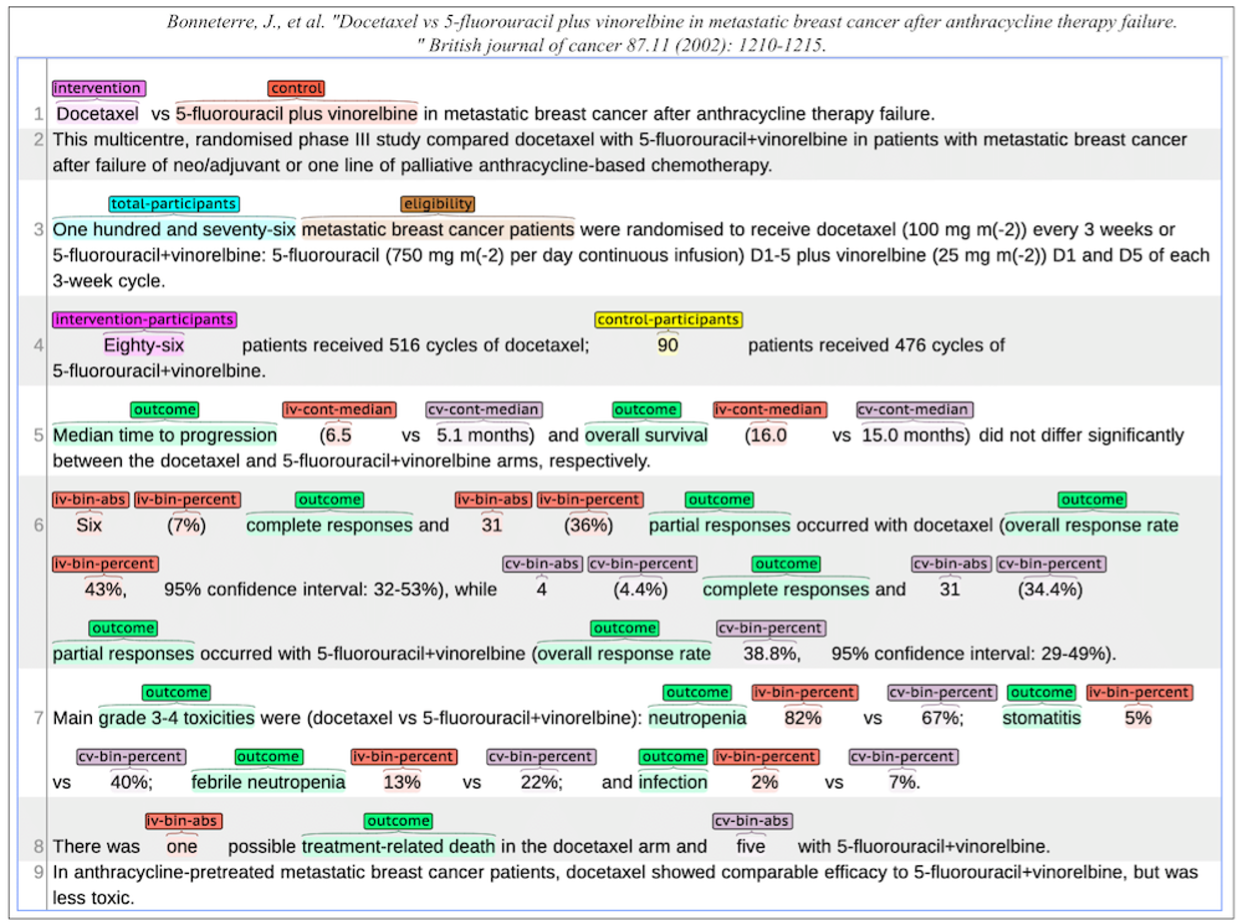}
\captionof{figure}{Annotation examples adopted from the PICO-Corpus (\url{https://github.com/sociocom/PICO-Corpus/tree/main}).}
\label{supfig:annotation_example}}

\end{document}